\documentclass{article}

\PassOptionsToPackage{numbers, compress, sort}{natbib}
\usepackage[final]{neurips_data_2021}

\usepackage[utf8]{inputenc} %
\usepackage[T1]{fontenc}    %
\usepackage[colorlinks=true]{hyperref}       %
\usepackage{url}            %
\usepackage{booktabs}       %
\usepackage{amsfonts}       %
\usepackage{amssymb}
\usepackage{mathtools}
\usepackage{nicefrac}       %
\usepackage{microtype}      %
\usepackage[svgnames]{xcolor}         %
\usepackage{graphicx}
\usepackage{subcaption}
\usepackage{multirow}
\usepackage{adjustbox}
\usepackage[capitalise]{cleveref}
\usepackage{wrapfig}
\usepackage{siunitx}
\usepackage{xspace}
\usepackage{makecell}
\usepackage{multicol}

\makeatletter
\newcommand{\ie}{\textit{i}.\textit{e}.\@ifnextchar{,}{}{~}}
\makeatother
\makeatletter
\newcommand{\eg}{\textit{e}.\textit{g}.\@ifnextchar{,}{}{~}}
\makeatother

\newcommand{\CLEVR}{\textsc{CLEVR}}
\newcommand{\CLEVRTEX}{\textsc{ClevrTex}}
\newcommand{\GRASSBG}{\textsc{GrassBG}}
\newcommand{\PBG}{\textsc{PlainBG}}
\newcommand{\VBG}{\textsc{VarBG}}
\newcommand{\TEST}{\textsc{OOD}} 
\newcommand{\CAMO}{\textsc{CAMO}}

\newcommand{\glimpse}{\(\mathrlap{\mathop{\hspace*{.15em}\raisebox{.15em}{\scalebox{.5}{$\Box$}}}}\Box\)}
\newcommand{\sprite}{\(\mathrlap{\mathop{\hspace*{.15em}\raisebox{.15em}{\scalebox{.5}{$\clubsuit$}}}}\Box\)}
\newcommand{\pixels}{\(\boxplus\)}

\title{\CLEVRTEX{}: A Texture-Rich Benchmark for Unsupervised Multi-Object Segmentation}

\author{
	Laurynas Karazija\\
    \And
	Iro Laina \\
    \And
	Christian Rupprecht \\
	\AND
	Visual Geometry Group, University of Oxford\\
	\texttt{\{laurynas, iro, chrisr\}@robots.ox.ac.uk}
}

\begin{document}

\maketitle

\begin{abstract}
There has been a recent surge in methods that aim to decompose and segment scenes into multiple objects in an unsupervised manner, \ie, unsupervised multi-object segmentation. 
Performing such a task is a long-standing goal of computer vision, offering to unlock object-level reasoning without requiring dense annotations to train segmentation models. 
Despite significant progress, 
current models are developed and trained on visually simple scenes depicting mono-colored objects on plain backgrounds. 
The natural world, however, is visually complex with confounding aspects such as diverse textures and complicated lighting effects. 
In this study, we present a new benchmark called \CLEVRTEX{}, designed as the next challenge to compare, evaluate and analyze algorithms. 
\CLEVRTEX{} features synthetic scenes with diverse shapes, textures and photo-mapped materials, created using physically based rendering techniques. It includes 50k examples depicting 3-10 objects arranged on a background, created using a catalog of 60 materials, and a further test set featuring 10k images created using 25 different materials.
We benchmark a large set of recent unsupervised multi-object segmentation models on \CLEVRTEX{} and find all state-of-the-art approaches fail to learn good representations in the textured setting, despite impressive performance on simpler data.
We also create variants of the \CLEVRTEX{} dataset, controlling for different aspects of scene complexity, and probe current approaches for individual shortcomings.  
Dataset and code are available at \url{https://www.robots.ox.ac.uk/~vgg/research/clevrtex}.
\end{abstract}

\section{Introduction}
 
Supervised scene understanding has seen significant progress in the last decade. 
The introduction of deep learning to the field and large, manually annotated datasets have made it possible to address tasks such as object detection~\cite{liu2020deep}, semantic or instance segmentation~\cite{he2017mask}, layout prediction~\cite{xu2017scene} and dense captioning~\cite{johnson2016densecap} with considerable accuracy. 
However, in absence of labels, and thereby supervision, such tasks are exceedingly difficult, even though it is easy to imagine that with enough images (or videos), it should be possible to identify objects and the general composition of a scene without human annotations.
This renders unsupervised multi-object segmentation, as well as object-centric learning a challenging yet promising field with high potential.

While certain tasks in the general context of \emph{unsupervised} scene understanding and decomposition have a relatively long history in computer vision, the majority of applications focus on single objects: 
image classification~\cite{van2020scan,Ji_2019_ICCV,caron2018deep}, saliency detection~\cite{zhang2018deep,nguyen2019deepusps}, foreground/background segmentation~\cite {chen2019unsupervised,bielski2019emergence,voynov2020big,melas2021finding} and general image-level representation learning~\cite{he2020momentum,chen2020simple,caron2021emerging,grill2020bootstrap}.
These methods are usually developed on datasets such as ImageNet~\cite{russakovsky2015imagenet} that contain one object of interest per image.
Nevertheless, most real-world scenes are often comprised of multiple objects in varying spatial configurations. 

\begin{figure}[t]
\centering
	\begin{subfigure}[l]{0.16\textwidth}
		\includegraphics[width=\textwidth, trim={0.0cm 1cm 0.0cm 1cm}, clip]{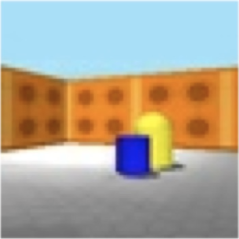}
		\caption*{\small GQN~\cite{eslami2018gqn}}
	\end{subfigure}
	\begin{subfigure}[l]{0.16\textwidth}
		\includegraphics[width=\textwidth, trim={0.0cm .4cm 0.0cm .15cm}, clip]{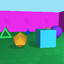}
		\caption*{\small ObjectRoom~\cite{burgess2019monet}}
	\end{subfigure}
	\begin{subfigure}[l]{0.16\textwidth}
		\includegraphics[width=\textwidth, trim={0.0cm .4cm 0.1cm 1.6cm}, clip]{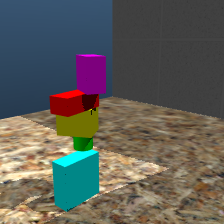}
		\caption*{\small ShapeStacks~\cite{groth2018shapestacks}}
	\end{subfigure}
	\begin{subfigure}[l]{0.16\textwidth}
		\includegraphics[width=0.98\textwidth, trim={0.3cm 0.0cm 0.3cm 0.2cm}]{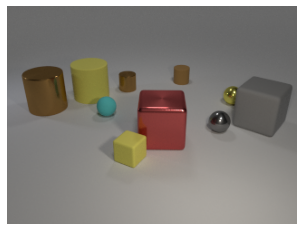}
	\caption*{\small \CLEVR{}~\cite{johnson2017clevr}}
	\end{subfigure}
	\begin{subfigure}[l]{.32\textwidth}
    	\begin{subfigure}[l]{0.49\textwidth}
    		\includegraphics[width=\textwidth, trim={0.3cm 0.0cm 0.3cm 0.2cm}]{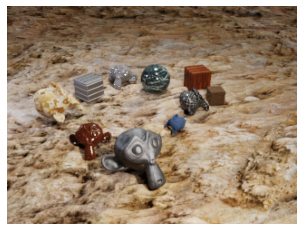}
    	\end{subfigure}
    	\begin{subfigure}[l]{0.49\textwidth}
    		\includegraphics[width=\textwidth, trim={0.3cm 0.0cm 0.3cm 0.2cm}]{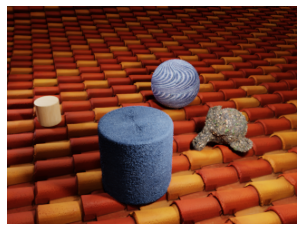}
    	\end{subfigure}
    	\caption*{\small \CLEVRTEX{}}
	\end{subfigure}
	\caption{Qualitative comparison of our new \CLEVRTEX{} dataset with previous unsupervised multi-object learning datasets featuring 3D objects. See \cref{tab:datasets} for quantitative comparison.}
	\label{fig:clevr_clevrtex_comparisor}
\end{figure}

Only recently, methods have been developed to analyze and decompose whole scenes containing multiple objects, \ie, jointly learning to represent and segment objects from raw image input, \emph{without} supervision. 
However, since moving from individual objects to complex scenes drastically complicates the problem, these methods currently rely on simple synthetic datasets. 
The complexity of these datasets ranges from simple, single-color 2D shapes arranged against a black background~\cite{burgess2019monet} to rendered 3D scenes composed of uniformly colored, 3D primitives 
(cubes, spheres, cylinders)~\cite{johnson2017clevr} (\cref{fig:clevr_clevrtex_comparisor}).
Interestingly, current methods work \emph{very well} on this kind of data and saturate the existing benchmarks such that a quantitative comparison of models becomes difficult.

How to scale such methods to visually complex real-world data remains an open problem.
When analyzing the current state-of-the-art methods and datasets, it becomes clear that there is a strong reliance on simple appearance (\eg, single color, simple shape).
For example, \citet{greff2019iodine} identify a tendency of their model to segment by color, and it fails when applied to natural images. 
In fact, the majority of methods learn semantic objects using similar compositional principles, which exploit statistical advantages in aligning simple scene elements with internal representations.
Natural images and the objects therein, however, do not possess strong, consistent colors. Instead, they feature confounding textures, often a mixture of repeating and irregular patterns, which might violate such assumptions.

This work introduces a dataset and benchmark as the next step towards eventually tackling real-world scenarios.
We propose \CLEVRTEX{}, a synthetic dataset that consists of \textit{textured} foreground objects and background, unlike existing benchmarks.
Interestingly, we find that simply moving from uniformly colored to textured objects poses extreme challenges for current models, and no existing method achieves satisfactory performance. 
For this reason, we also introduce several variants of our dataset to gradually scale the visual complexity of the scenes and investigate where current algorithms struggle. 
To probe the generalization capability of models to out-of-distribution scenes, we create additional test sets that contain unseen shapes and materials and camouflaged objects. 
Together with \CLEVRTEX{} and its variants, we are releasing the code to generate the dataset from scratch.
Finally, we find that existing work does not rely on a consistent set of metrics and benchmarks.
In an extensive set of experiments, we benchmark the majority\footnote{wherever code was available or could be obtained from the authors} of current work on both \CLEVR{} and our newly introduced \CLEVRTEX{}.
\begin{table}
\caption{Comparison of the proposed \CLEVRTEX{} dataset with previous unsupervised multi-object learning datasets featuring 3D objects. 
}
\label{tab:datasets}
\centering
\scalebox{.72}{
\begin{tabular}{lrrrrrrl}
\toprule
\multirow{2}{*}{Dataset} & \multirow{2}{*}{\#Images} & \multirow{2}{*}{\#Objects} & \multirow{2}{*}{\#Shapes} & \#Obj. & \#Obj. & \multirow{2}{*}{\#Backgrounds} & \multirow{2}{*}{Annotations} \\
& & & & Colors & Materials & & \\
\midrule
GQN~\citep{eslami2018gqn} & 12M & 1--3 & 7 & --- & 1 & 15 & Camera parameters \\
ObjectRoom~\citep{burgess2019monet} & 1M & 1--6 & 4 & 10 & 1 & 100 & Semantic, factor of variation \\
ShapeStacks~\citep{groth2018shapestacks} & 310k & 2--6 & 4 & 5 & 1 & 25 & Semantic, stability, stability type\\
\CLEVR{}~\citep{johnson2017clevr} & 100k & 3--10 & 3 & 8 & 2 & 1 & Semantic, factors of variation \\
\CLEVRTEX{} (Ours) & 50k+10k & 3--10 & 4+4 & --- & 60+25 & 60+25 & \makecell[l]{Semantic, depth, normal, shadow,\\factors of variation} \\
\bottomrule
\end{tabular}
}
\end{table}

\section{Related Work}\label{sec:rel_work}

Object recognition benchmarks such as PascalVOC~\cite{everingham2010pascal} or MS COCO~\cite{lin2014microsoft} have been fundamental to object detection research. 
However, the current unsupervised multi-object segmentation models are yet unable to handle diverse real-world images featured in such datasets and have relied on visually trivial 2D and 3D data. Here, we review datasets and benchmarks used in \textit{unsupervised} multi-object segmentation methods and point out the differences to \CLEVRTEX{}. 

\paragraph{2D Datasets}
Earlier unsupervised multi-object learning approaches were applied to transformed versions of existing 2D datasets, often originally crafted for disentanglement research, such as Shapes~\cite{reichert2011hierarchical}, variants of MNIST~\citep{lecun1998gradient}: TexturedMNIST~\cite{greff2016tagger} and MultiMNIST~\cite{sabour2017dynamic}, as well as the multi-object version of dSprites~\cite{dsprites17}, \ie, Multi-dSprites~\cite{burgess2019monet}. 
Others borrow data from the reinforcement learning community, such as the ATARI game environment~\citep{bellemare2013arcade} or Tetrominoes~\cite{tetrominoes19}. 
However, 2D datasets, whilst valuable for development, do not contain the visual cues and details (\eg shadows and perspective) needed to learn object segmentation that generalizes to real images.

\paragraph{3D Datasets} 
Simple 3D Phong-shaded datasets (\cref{fig:clevr_clevrtex_comparisor}) have been crafted for use in the unsupervised multi-object setting. 
The object-room dataset~\cite{burgess2019monet}, a multi-object extension of 3D shapes~\cite{3dshapes18}, features colored shapes arranged in a room with colored walls.
ShapeStacks~\citep{groth2018shapestacks} features stacked, solid-colored primitives on a simple background with a pattern.
\CLEVR{}~\cite{johnson2017clevr}, which is most closely related to our work, was introduced as a visual question-answering dataset but has become a popular benchmark in unsupervised scene decomposition as well. 
It features a set of 3-10 primitive shapes arranged on a gray photo backdrop; objects can have either a rubbery or metallic appearance and one of 8 color tints. 
\CLEVR6~\cite{greff2019iodine} is a filtered version of the \CLEVR{} dataset that includes only up to 6 objects per image.
It is often used for training in multi-object representation learning, with the remainder of \CLEVR{} used to test generalization to more crowded scenes \cite{locatello2020object, emami2021efficient}.

Additional variants of \CLEVR{} have also been generated for other tasks, such as ARROW~\citep{jiang2020generative} for exploring scene composition accuracy, and a recursive version in \cite{deng2021generative} for learning part-whole relationships. 
Multi-view variations~\cite{kosiorek2021nerf, stelzner2021decomposing} are used for 3D representation learning, and further include new object geometry, such as toys \cite{nanbo2020learning} and chairs \cite{yu2021unsupervised}. 
However, these datasets feature simple scenes of low visual complexity, with contrasting solid colors present on objects. \CLEVRTEX{} instead contains difficult objects with various materials that include repeating patterns and small details and often blend in rather than stand out from the background.

The main differences in data statistics between \CLEVRTEX{} and commonly used multi-object learning datasets are also summarised in \cref{tab:datasets}.

\paragraph{Unsupervised Multi-Object Segmentation in Natural Scenes} 
Some attempts have also been made to scale to natural scenes. 
\citet{eslami2016attend} apply the AIR model modified with a 3D rendering engine to infer identities and positions of crockery items on a table, training on simulated data, and evaluating against real-world images. 
\citet{monnier2021dtisprites} test their sprite-based method on foreground/background segmentation on the Weizmann Horse dataset~\citep{borenstein2004learning}.
\citet{engelcke2021genesis} apply Genesis-V2  to robotic manipulation datasets, Sketchy and APC \citep{zeng2017multi}.
Sketchy \citep{cabi2019scaling} features recordings of a robotic arm manipulating solid colored toys, towels, or other small objects on a test table, but it lacks segmentation masks. 
The APC \citep{zeng2017multi} dataset is used instead for evaluation but only contains a single foreground object. 
These attempts signal promise that unsupervised multi-object segmentation can eventually scale to diverse real-world images.

\paragraph{Visual Fidelity in Simulation} 
Simulation has always been central to progress in machine/reinforcement learning. 
However, as usual, the gap between a simulated setting and the ability to generalize to real-world environments is of concern. 
Several new simulators aim to improve the visual fidelity using photo-mapped environments or artists' compositions~\citep{savva2017minos,kolve2017ai2,xia2018gibson,habitat19iccv}.
Recently, TWD~\citep{gan2021threedworld} introduced a rich physics engine and PBR rendering of environments with a library of objects. 
Similar to our work, the emphasis is partly on increasing visual fidelity while moving away from trivial settings and towards real-world applications. 
However, RL environments have not seen much use in the unsupervised vision domain due to the often specific nature of the data, egocentric perspective, and temporal dependency. 

\section{\CLEVRTEX}
We introduce \CLEVRTEX{}, a simulated dataset designed to present the next challenge in unsupervised multi-object learning. 
It introduces confounding visual aspects such as texture, irregular shapes, and various materials while maintaining control over scene composition. \CLEVRTEX{} is available under CC-BY license.

\subsection{Dataset Creation}
\label{sec:creation}

\CLEVRTEX{} is a much more visually complex extension of \CLEVR{}~\citep{johnson2017clevr} targeted at multi-object learning.
It is procedurally generated using the API of Blender\footnote{\url{https://www.blender.org/}}, a powerful open-source 3D suite.

At the center of the \CLEVRTEX{} generation process is a catalog of diverse photo-mapped materials\footnote{We use the computer graphics term ``material'' to refer to the collection of resources used to creates the likeness of appropriate real-world material on simulated surfaces. Materials are typically a composition of various modalities, such as normal, diffuse, specular, and displacement maps, as well as a computation graph and shaders. We use the term ``texture'' to refer to 2D images mapping color information onto 3D surfaces.} ranging from forest floor duff, rocks, brickwork, and tiles to fabrics, metallic weaves, and meshes\,---\,a full list of materials is shown in ~\cref{sec:dataset_construction}).
To generate each image, we start with a scene containing only a photo backdrop, which will become the background. 
For viewpoint and lighting diversity, we apply random jitter to the position of the camera and three lights. 
We then fill the scene with 3 to 10 objects (number sampled uniformly),
sampling each object from a set of shapes: a cube, a sphere, a cylinder, and a non-symmetric shape of anthropomorphized monkey head\footnote{A modified version of Suzzane -- a prefab shape available in Blender.} for increased complexity in object silhouettes. 
Objects are added to the scene one by one by sampling position (continuous,  $ (x, y) \sim \mbox{Uniform}(-3, 3)$), scale (discrete, $s \in \{.9, .6, .4\}$), and rotation (continuous, $\theta \sim \mbox{Uniform}(0, 360)$).
If a new object collides with already existing shapes in the scene, the object's transformation is resampled until no collision is found or a maximum number of trials is exceeded, at which the process restarts by removing all objects.

\begin{wrapfigure}{r}{.42\textwidth}
\vspace{-1.65em}
    \begin{flushright} 
        \rotatebox[origin=tl]{90}{\hspace*{-.7em}\tiny \CLEVRTEX{}} 
    	\begin{subfigure}[t]{.4\textwidth}
    	\centering
    	\begin{subfigure}[t]{0.23\textwidth}
    		\includegraphics[width=\textwidth, trim={0.3cm 1cm 0.3cm 0.2cm}]{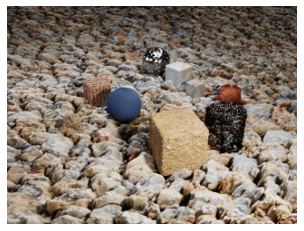}
    	\end{subfigure}
    	\begin{subfigure}[t]{0.23\textwidth}
    		\includegraphics[width=\textwidth, trim={0.3cm 1cm 0.3cm 0.2cm}]{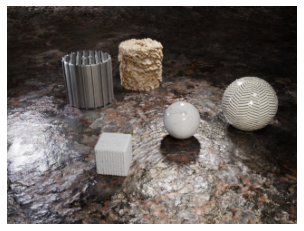}
    	\end{subfigure}
    	\begin{subfigure}[t]{0.23\textwidth}
    		\includegraphics[width=\textwidth, trim={0.3cm 1cm 0.3cm 0.2cm}]{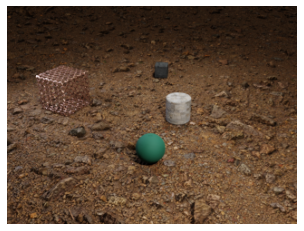}
    	\end{subfigure}
    	\begin{subfigure}[t]{0.23\textwidth}
    		\includegraphics[width=\textwidth, trim={0.3cm 1cm 0.3cm 0.2cm}]{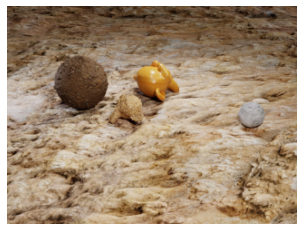}
    	\end{subfigure}
    	\end{subfigure}
    \end{flushright}
    \begin{flushright} 
        \rotatebox[origin=tl]{90}{\hspace*{.1em}\tiny \TEST{}} 
    	\begin{subfigure}[t]{.4\textwidth}
    	\centering
    	\begin{subfigure}[t]{0.23\textwidth}
    		\includegraphics[width=\textwidth, trim={0.3cm 1cm 0.3cm 0.2cm}]{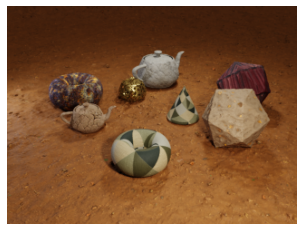}
    	\end{subfigure}
    	\begin{subfigure}[t]{0.23\textwidth}
    		\includegraphics[width=\textwidth, trim={0.3cm 1cm 0.3cm 0.2cm}]{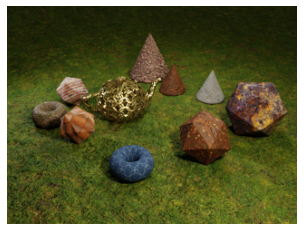}
    	\end{subfigure}
    	\begin{subfigure}[t]{0.23\textwidth}
    		\includegraphics[width=\textwidth, trim={0.3cm 1cm 0.3cm 0.2cm}]{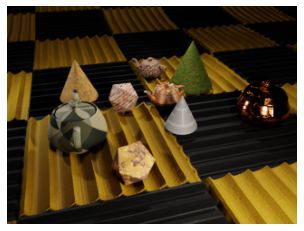}
    	\end{subfigure}
    	\begin{subfigure}[t]{0.23\textwidth}
    		\includegraphics[width=\textwidth, trim={0.3cm 1cm 0.3cm 0.2cm}]{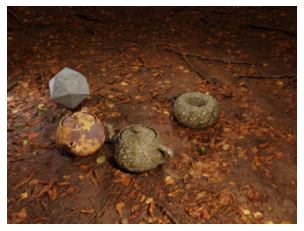}
    	\end{subfigure}
    	\end{subfigure}
    \end{flushright}
    \begin{flushright} 
            \rotatebox[origin=tl]{90}{\hspace*{.0em}\tiny \CAMO{}} 
    	\begin{subfigure}[t]{.4\textwidth}
    	\centering
    	\begin{subfigure}[t]{0.23\textwidth}
    		\includegraphics[width=\textwidth, trim={0.3cm 0.5cm 0.3cm 0.2cm}]{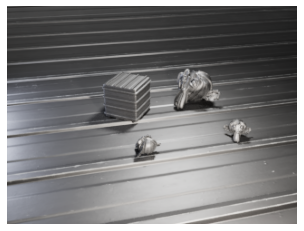}
    	\end{subfigure}
    	\begin{subfigure}[t]{0.23\textwidth}
    		\includegraphics[width=\textwidth, trim={0.3cm 0.5cm 0.3cm 0.2cm}]{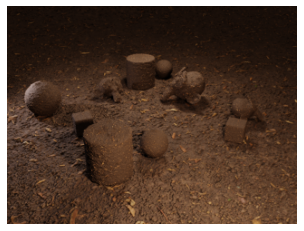}
    	\end{subfigure}
    	\begin{subfigure}[t]{0.23\textwidth}
    		\includegraphics[width=\textwidth, trim={0.3cm 0.5cm 0.3cm 0.2cm}]{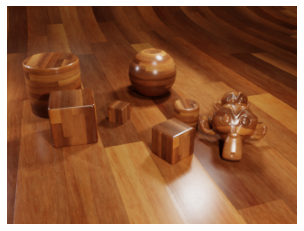}
    	\end{subfigure}
    	\begin{subfigure}[t]{0.23\textwidth}
    		\includegraphics[width=\textwidth, trim={0.3cm 0.5cm 0.3cm 0.2cm}]{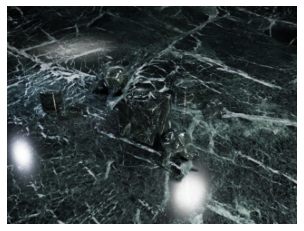}
    	\end{subfigure}
    	\end{subfigure}
    \end{flushright}
    \begin{flushright} 
        \rotatebox[origin=tl]{90}{\hspace*{-.6em}\tiny \PBG{}} 
    	\begin{subfigure}[t]{.4\textwidth}
    	\centering
    	\begin{subfigure}[t]{0.23\textwidth}
    		\includegraphics[width=\textwidth, trim={0.3cm 1cm 0.3cm 0.2cm}]{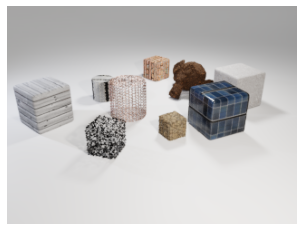}
    	\end{subfigure}
    	\begin{subfigure}[t]{0.23\textwidth}
    		\includegraphics[width=\textwidth, trim={0.3cm 1cm 0.3cm 0.2cm}]{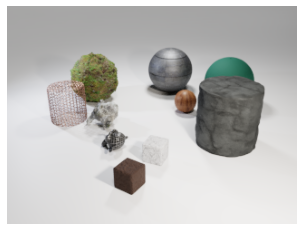}
    	\end{subfigure}
    	\begin{subfigure}[t]{0.23\textwidth}
    		\includegraphics[width=\textwidth, trim={0.3cm 1cm 0.3cm 0.2cm}]{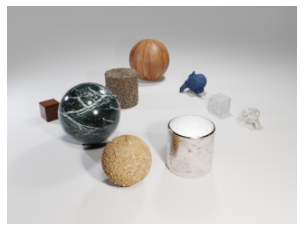}
    	\end{subfigure}
    	\begin{subfigure}[t]{0.23\textwidth}
    		\includegraphics[width=\textwidth, trim={0.3cm 1cm 0.3cm 0.2cm}]{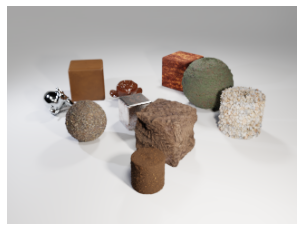}
    	\end{subfigure}

	    \end{subfigure}
	\end{flushright}
	\begin{flushright} 
        \rotatebox[origin=tl]{90}{\hspace*{-.4em}\tiny \VBG{}} 
    	\begin{subfigure}[t]{.4\textwidth}
    	\centering
    	\begin{subfigure}[t]{0.23\textwidth}
    		\includegraphics[width=\textwidth, trim={0.3cm 1cm 0.3cm 0.2cm}]{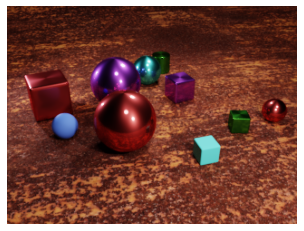}
    	\end{subfigure}
    	\begin{subfigure}[t]{0.23\textwidth}
    		\includegraphics[width=\textwidth, trim={0.3cm 1cm 0.3cm 0.2cm}]{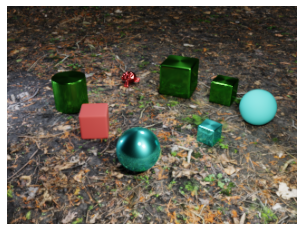}
    	\end{subfigure}
    	\begin{subfigure}[t]{0.23\textwidth}
    		\includegraphics[width=\textwidth, trim={0.3cm 1cm 0.3cm 0.2cm}]{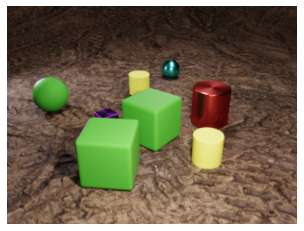}
    	\end{subfigure}
    	\begin{subfigure}[t]{0.23\textwidth}
    		\includegraphics[width=\textwidth, trim={0.3cm 1cm 0.3cm 0.2cm}]{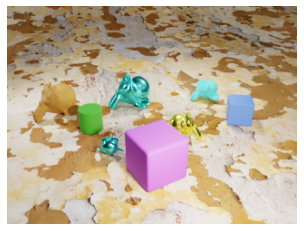}
    	\end{subfigure}
    	\end{subfigure}
	\end{flushright}
	\begin{flushright} 
        \rotatebox[origin=tl]{90}{\hspace*{-.5em}\tiny \GRASSBG{}} 
    	\begin{subfigure}[t]{.4\textwidth}
    	\centering
    	\begin{subfigure}[t]{0.23\textwidth}
    		\includegraphics[width=\textwidth, trim={0.3cm 1cm 0.3cm 0.2cm}]{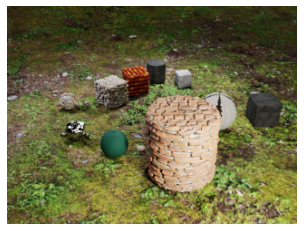}
    	\end{subfigure}
    	\begin{subfigure}[t]{0.23\textwidth}
    		\includegraphics[width=\textwidth, trim={0.3cm 1cm 0.3cm 0.2cm}]{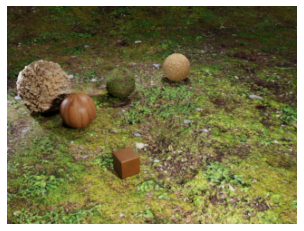}
    	\end{subfigure}
    	\begin{subfigure}[t]{0.23\textwidth}
    		\includegraphics[width=\textwidth, trim={0.3cm 1cm 0.3cm 0.2cm}]{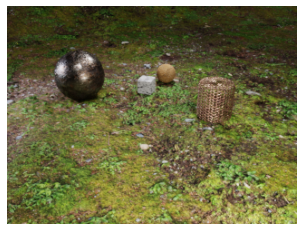}
    	\end{subfigure}
    	\begin{subfigure}[t]{0.23\textwidth}
    		\includegraphics[width=\textwidth, trim={0.3cm 1cm 0.3cm 0.2cm}]{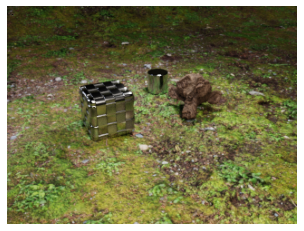}
    	\end{subfigure}
    	\end{subfigure}
	\end{flushright}
	\caption{\CLEVRTEX{} and its variants.}
	\label{fig:variants}
\vspace{-1.5em}
\end{wrapfigure}
We then sample a material for each object and the background.
Using adaptive subdivision, we create material-specific geometry by displacing vertices of the starting shapes. 
This creates reliefs for simpler materials or distorts shapes, extruding features or introducing holes. 
The materials use albedo, subsurface scattering, and reflectivity maps to generate detailed visuals. 
Using physically based rendering ensures appropriately detailed reflections, highlights, and lighting effects. 
In addition, we generate ground truth segmentation maps through the rendering process and automatically check that no object is fully occluded.
In that case, the scene is resampled from scratch. Further figures depicting scene lighting, objects, their scales and deformations are available in \cref{sec:dataset_construction}.

The object shapes and placement mimics that of the \CLEVR{} dataset \citep{johnson2017clevr} for backward compatibility. 
We do not generate the question-answering part of the original \CLEVR{} dataset but include full metadata.
This means that this dataset could also be used for other \CLEVR-based tasks such as question answering, although this is not our focus here.
Similarly, in anticipation that our dataset might also find usages beyond its intended setting, we include depth, albedo, shadow, and normal maps alongside the images, segmentation maps, and metadata. 
We share the code to generate \CLEVRTEX{} alongside the dataset.

\subsection{Statistics}

\CLEVRTEX{} contains $50\,000$ images, of which we use 10\% for testing,  10\% for validation and the remaining 80\% ($40\,000$ images) for training. 
Each image contains between three and ten objects (uniformly sampled). 
There are four possible shapes, which have been modified to enable clean texture mapping. 
We use three distinct object scales to maintain identifiable size ``names'', as in \CLEVR{}, and custom meshes to ensure that the scaling of the objects does not distort texture details. 
The object placement and rotation are sampled from a continuous range. 
Note that even though two shapes\,---\,cylinder and sphere\,---\,are rotationally symmetric, the materials applied to them are not. 
We use a catalog of 60 materials with non-commercial licenses to generate the whole dataset before splitting the data into training sets. 
The materials are manually adjusted to ensure visually pleasing results at different scales and the background. 

\subsection{Variants}
We create the following modifications of \CLEVRTEX{}, each with $20\,000$ images (see \cref{fig:variants}), to enable a more detailed analysis and evaluation and probe methods for their shortcomings.

The first variant, \PBG{}, is a dataset consisting of textured objects on a plain background, \ie, the background is always set to a simple material as in \CLEVR{}. 
We also create the reverse version, \VBG{} (varied background), where the objects are assigned simple \CLEVR-like materials and colors while the background receives a textured material at random from our material catalog.
\PBG{} and \VBG{} fall in-between \CLEVR{} and \CLEVRTEX{} in terms of visual complexity.  
In \PBG{}, intra-object appearance is more complex, but each object clearly stands out from the plain background. 
On the other hand, \VBG{} maintains uniformly colored objects but introduces background texture, effectively making the background more diverse than the foreground.
\PBG{} and \VBG{} can be used to analyze the importance of background vs.~object reconstruction.

Furthermore, we create \GRASSBG{}, which contains scenes with the same mossy grass material as the background, while foreground objects receive materials at random.
This variant is thus comparable to \CLEVRTEX{} in terms of visual complexity. 
However, consistency in the background allows for testing memorization vs.~reconstruction effects.

In addition, we propose the following two test sets to serve as an extra check for the limitations of \CLEVRTEX{}. 

\CAMO{} contains scenes with ``camouflaged'' objects.
To simulate this, every scene is made of a single, randomly sampled material that is used on all objects \textit{and} the background.
\CAMO{} is created to challenge the internal-vs-external consistency and the efficiency hypothesis that underpins compositional methods.
The only visual cues here are lighting, shadows, and perspective. 
It should enable probing models to see if they rely on such context to identify objects.
Although we release \CAMO{} with training, validation and test splits, in our experiments it is only used as a testbed for models trained on \CLEVRTEX. 

Finally, we also provide a separate \TEST{} (out-of-distribution) dataset to evaluate model generalization on novel scenes. 
This dataset is designed exclusively as a test set and thus only contains $10\,000$ images.
\TEST{} is generated the same way as \CLEVRTEX{}, but exclusively uses 25 \emph{new} (unseen) materials\,---\,\ie different from the 60 already used in other variants\,---\,and four new shapes (cone, torus, icosahedron, and a teapot) that are not part of \CLEVRTEX{}.
\section{Models}
\label{sec:models}
In recent years, there has been a surge of methods that aim to decompose a scene into objects in an unsupervised manner and, at the same time, learn object-centric representations.  
Following \cite{lin2020space}, we categorize these methods as follows. 

\paragraph{Pixel-Space Approaches (\pixels{})}
A common way to frame the problem of unsupervised scene decomposition into objects is to assign each pixel to one of a usually fixed number of scene components, inferring per-pixel membership maps \cite{greff2016tagger,greff2017neural,greff2019iodine,burgess2019monet,yang2020learning, emami2021efficient}. 
While these methods are probabilistic in nature, they do not lend themselves to generating new images.
For this reason, several generative methods have been proposed, where images can be sampled from the learned distributions \cite{engelcke2019genesis,engelcke2021genesis}.
Finally, \citet{locatello2020object} introduce a discriminative approach using an iterative clustering-like slot attention mechanism.

Here, we benchmark MONet~\citep{burgess2019monet} and IODINE~\citep{greff2019iodine} as examples of earlier approaches that handle 3D colored scenes. We also evaluate the improved efficient MORL (eMORL)~\citep{emami2021efficient},  Genesis-v2~\cite{engelcke2021genesis} as a generative model, and Slot Attention~\citep{locatello2020object} which is representative for discriminative models.

\paragraph{Glimpse-Based Methods (\glimpse{})}
An alternative to predicting components for each pixel is to extract patches of the input---named \textit{glimpses}---that contain objects. 
A dense segmentation can be derived in this reduced space. 
These glimpses are arranged on top of an explicit background to reconstruct the image. 
Glimpse-based methods~\cite{eslami2016attend,crawford2019spatially,lin2020space,jiang2020generative,deng2021generative} tend to offer computational advantages due to smaller regions, however also require deciding, extracting and composing patches.

\newcommand{\itime}[2]{
\hspace{-1.5em}\tablenum[table-format = 2.0]{#1}\hspace{-2em}\small\color{DarkSlateBlue}$\pm#2$
}
\begin{wraptable}{r}{.52\textwidth}
\vspace{-0.1em}
\caption{
Computational resources for different models.
$\times$ indicates number of GPUs needed. 
Measured on NVIDIA P40 24GB GPUs, with original batch sizes and $128 \times 128$ input. 
Train. time refers to time required to train the models for the recommended number of iterations, measured in total GPU hours. Inf. time measures the mean inference time required for a single batch, shown $\pm \sigma$ over 7 passes.
}
\label{tab:comp_res}
\centering
\scalebox{.85}{
\begin{tabular}{lccc}
\toprule
\multirow{2}{*}{Model} & Train. Time & Inf. Time & Peak GPU  \\
& (GPU h) & (ms{\small\color{DarkSlateBlue}$\pm\sigma$}) & Mem (GB) \\
\midrule
\glimpse{} GNM \citep{jiang2020generative} & 54 & \itime{258}{9} & 4 \\
\glimpse{} SPACE \citep{lin2020space} & 64 & \itime{191}{2} & 8 \\
\glimpse{} SPAIR* \citep{crawford2019spatially} & 77 & \itime{213}{2} & 11 \\
\sprite{} DTI \citep{monnier2021dtisprites} & 198 & \itime{2530}{5} & 11 \\
\sprite{} MN \citep{smirnov2021marionette} & --- & --- & 11 \\
\pixels{} IODINE \citep{greff2019iodine} & \(4 \times 202\) & \itime{1360}{2} & \(4 \times 23\) \\
\pixels{} SA \citep{locatello2020object} & 290 & \itime{818}{1} & 17 \\
\pixels{} MONet \citep{burgess2019monet} & \(3 \times 106\) & \itime{544}{1} & \(3 \times 17\) \\
\pixels{} eMORL \citep{emami2021efficient} & \(4 \times 158\) & \itime{217}{1} & \(4 \times 17\) \\
\pixels{} GenV2 \citep{engelcke2021genesis} & 194 & \itime{452}{1} & 15 \\
\bottomrule
\end{tabular}
}
\vspace{-2.6em}
\end{wraptable}

From the glimpse-based methods, we benchmark SPAIR~\citep{crawford2019spatially}, which models glimpses auto-regressively, using a truncated geometric prior. 
Since it cannot handle non-black backgrounds, we modify the model to include a VAE for background prediction (SPAIR*). 
We also evaluate SPACE~\citep{lin2020space} due to its use of the pixel-space approach for processing the background, and GNM~\citep{jiang2020generative}, which uses scene-level priors.

\paragraph{Sprite-Based Methods (\sprite{})}
Recently, several methods~\citep{smirnov2021marionette,monnier2021dtisprites} propose to decompose images into a learned dictionary of RGBA sprites instead of learning a generative model. 
From the alpha masks of each sprite, the scene segmentation can be recovered. 
We benchmark MarioNette~\citep{smirnov2021marionette} and DTISprites~\citep{monnier2021dtisprites} to investigate the differences of two sprite-based (\sprite{}) approaches.

The aforementioned models have highly varying computational requirements. 
We offer a side-by-side comparison in \cref{tab:comp_res}, where computational advantages to glimpse-based methods can be immediately seen, with methods such as GNM and SPACE taking a fraction of time and memory required by even single-GPU pixel-space methods. 
All implementation details, hyper-parameters, and model changes are reported in \cref{sec:hyperparameters}.
\section{Experiments}

\paragraph{Datasets}
We benchmark a wide spectrum of methods using \CLEVRTEX{} and its variants.
To test generalization, we evaluate models trained on \CLEVRTEX{}  using \TEST{} and \CAMO{}.
In addition to our \CLEVRTEX{} and its variants, we conduct experiments on \CLEVR{} to provide a complete side-by-side comparison of methods and the new challenges in \CLEVRTEX{}.
All implementation details and preprocessing are reported in \cref{sec:sup_data}.

\paragraph{Metrics}
The majority of previous work has used the adjusted Rand index on foreground pixels (ARI-FG) only as an evaluation metric. 
We share concerns with \cite{monnier2021dtisprites, engelcke2019genesis} that this metric does not reflect how well objects are localized by the model and whether they are considered part of the background.
Thus, we report mean intersection over union (mIoU) instead, as it considers the background. 
Further discussion and a side-by-side comparison of ARI-FG and mIoU can be found in \cref{sec:sup_metrics}.
Furthermore, we judge the quality of the reconstruction output of the models using the mean squared error (MSE). 
For the models trained on \CLEVR{} and \CLEVRTEX{}, we report results on three random seeds, including their standard deviation. 
\begin{figure}[!ht]
    \includegraphics[width=.95\textwidth, trim={0 5 26 0}]{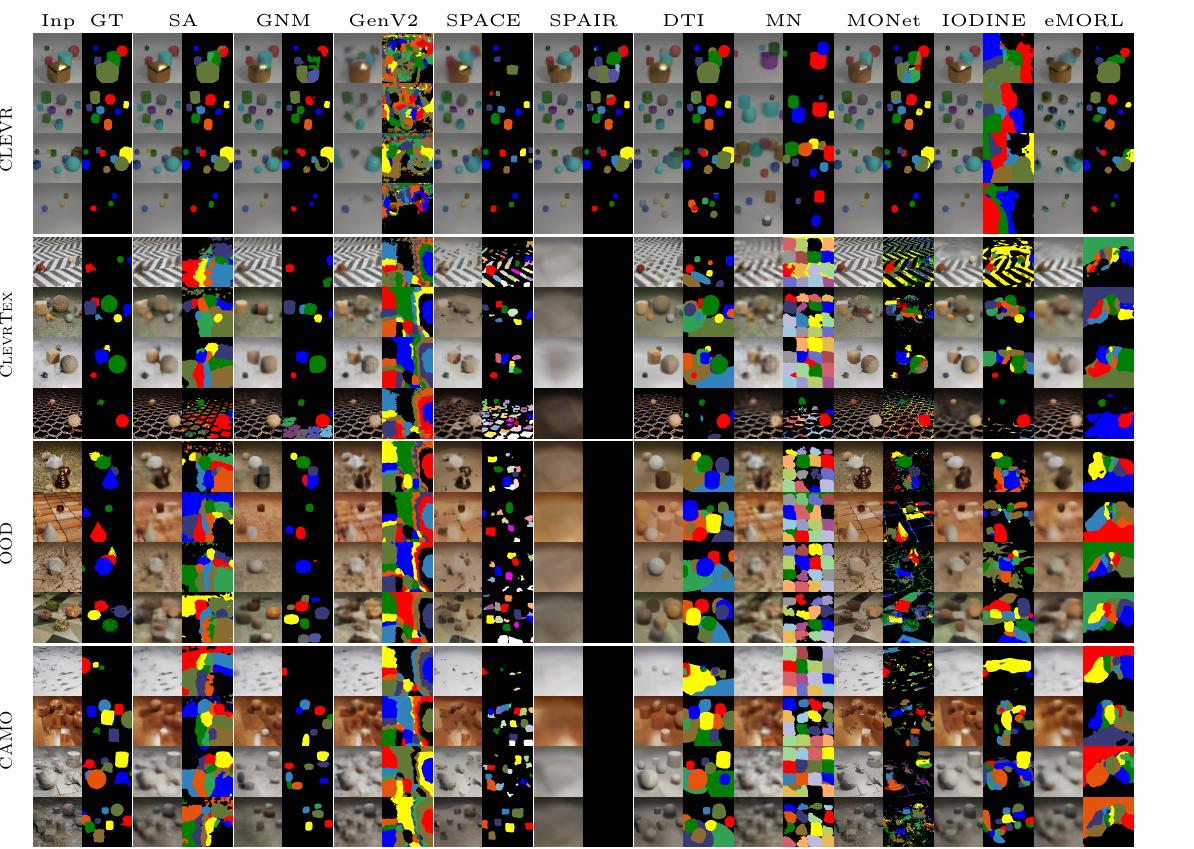}
	\caption{Comparison of various models' reconstruction and segmentation outputs on \CLEVR{}, \CLEVRTEX{} and our test sets. Best viewed digitally. More results in the Appendix,~\cref{fig:output_examples}.}
	\label{fig:output_examples_main}
\end{figure}
\subsection{Benchmark}
\newcommand{\miou}[2]{
\hspace{-.5em}\tablenum[table-format = 2.2]{#1}\small\color{DarkSlateBlue}$\pm$\tablenum[table-format = 2.2]{#2}
}
\newcommand{\mse}[2]{
\tablenum[table-format = 2.0]{#1}\small\color{DarkSlateBlue}$\pm$\tablenum[table-format = 3.0]{#2}\hspace{-1em}
}
\newcommand{\mioup}[1]{
\hspace{-.4em}\tablenum[table-format = 2.2]{#1}\hspace*{3.13em}
}
\newcommand{\msep}[1]{
\tablenum[table-format = 2.0]{#1}\hspace*{1.3em}
}
\begin{table}
\caption{Benchmark results on \CLEVR{} and \CLEVRTEX{} and the generalization test sets \CAMO{}, and \TEST{}. Results shown \(\pm \sigma\) calculated over 3 runs. $\dagger$ updated eMORL: after \CLEVRTEX{} was released, the authors of \citep{emami2021efficient} have updated their codebase to include \CLEVRTEX{} training and evaluation and shared their trained models with improved performance (single seed on \CLEVR{}).}
\label{tab:results_clevr}
\centering
\scalebox{.78}{
\begin{tabular}{lrr l rr l rr l rr}
\toprule
\multirow{2}{*}{Model} & \multicolumn{2}{c}{\CLEVR{}} 
& 
      & \multicolumn{2}{c}{\CLEVRTEX{}} 
&   
      & \multicolumn{2}{c}{\TEST{}}
&   
      & \multicolumn{2}{c}{\CAMO{}}
\\
& \(\uparrow\)mIoU (\%) 
& \(\downarrow\)MSE 
& 
& \(\uparrow\)mIoU (\%) 
& \(\downarrow\)MSE 
& 
& \(\uparrow\)mIoU (\%) 
& \(\downarrow\)MSE 
&
& \(\uparrow\)mIoU (\%) 
& \(\downarrow\)MSE 
\\
\cmidrule[\lightrulewidth]{1-3}
\cmidrule[\lightrulewidth]{5-6}
\cmidrule[\lightrulewidth]{8-9}
\cmidrule[\lightrulewidth]{11-12}

\glimpse{} SPAIR* \citep{crawford2019spatially}
& \miou{65.95}{4.02} & \mse{55}{10}
&
& \miou{0.0}{0.0} & \mse{1101}{2}
&
& \miou{0.0}{0.0} & \mse{1166}{5}
&
& \miou{0.0}{0.0} & \mse{668}{3}
\\

\glimpse{} SPACE \citep{lin2020space} 
& \miou{26.31}{12.93} & \mse{63}{3}
&
& \miou{9.14}{3.46} & \mse{298}{80}
&
& \miou{6.87}{3.32} & \mse{387}{66}
&
& \miou{8.67}{3.50} & \mse{251}{61}
\\

\glimpse{} GNM \citep{jiang2020generative} 
& \miou{59.92}{3.72} & \mse{43}{3}
&
& \miou{42.25}{0.18} & \mse{383}{2}
&
& \miou{40.84}{0.30} & \mse{626}{5}
&
& \miou{17.56}{0.74} & \mse{353}{1}
\\

\sprite{} MN \citep{smirnov2021marionette} 
& \miou{56.81}{0.40} & \mse{75}{1}
&
& \miou{10.46}{0.10} & \mse{335}{1}
&
& \miou{12.13}{0.19} & \mse{409}{3}
&
& \miou{8.79}{0.15} & \mse{265}{1}
\\

\sprite{} DTI \citep{monnier2021dtisprites} 
& \miou{48.74}{2.17} & \mse{77}{12}
&
& \miou{33.79}{1.30} & \mse{438}{22}
&
& \miou{32.55}{1.08} & \mse{590}{4}
&
& \miou{27.54}{1.55} & \mse{377}{17}
\\

\pixels{} GenV2 \citep{engelcke2021genesis} 
& \miou{9.48}{0.55} & \mse{158}{2}
&
& \miou{7.93}{1.53} & \mse{315}{106}
&
& \miou{8.74}{1.64} & \mse{539}{147}
&
& \miou{7.49}{1.67} & \mse{278}{75}
\\

\pixels{} eMORL \citep{emami2021efficient}
& \miou{50.19}{22.56} & \mse{33}{8}
&
& \miou{12.58}{2.39} & \mse{318}{43}
&
& \miou{13.17}{2.58} & \mse{471}{51}
&
& \miou{11.56}{2.09} & \mse{269}{31}
\\

\pixels{} eMORL$^\dagger$ \citep{emami2021efficient} 
& \mioup{21.98} & \msep{26}
&
& \miou{30.17}{2.60} & \mse{347}{20}
&
& \miou{25.03}{1.99} & \mse{546}{4}
&
& \miou{19.13}{4.88} & \mse{315}{21}
\\

\pixels{} MONet \citep{burgess2019monet}
& \miou{30.66}{14.87} & \mse{58}{12}
&
& \miou{19.78}{1.02} & \mse{146}{7}
&
& \miou{19.30}{0.37} & \mse{231}{7}
&
& \miou{10.52}{0.38} & \mse{112}{7}
\\

\pixels{} SA \citep{locatello2020object}
& \miou{36.61}{24.83} & \mse{23}{3}
&
& \miou{22.58}{2.07} & \mse{254}{8}
&
& \miou{20.98}{1.59} & \mse{487}{16}
&
& \miou{19.83}{1.41} & \mse{215}{7}
\\

\pixels{} IODINE \citep{greff2019iodine}
& \miou{45.14}{17.85} & \mse{44}{9}
&
& \miou{29.16}{0.75} & \mse{340}{3}
&
& \miou{26.28}{0.85} & \mse{504}{3}
&
& \miou{17.52}{0.75} & \mse{315}{3}
\\

\bottomrule
\end{tabular}
}
\end{table}

The results for the benchmark are detailed in \cref{tab:results_clevr} and in \cref{fig:output_examples_main}.
Next, we discuss our findings regarding the ability of models to separate foreground and background, to handle textured scenes, as well as their training stability and generalizability to new scenes.

\paragraph{Background Segmentation}
Pixel-space methods (\pixels{}) show impressive performance on \CLEVR{} compared against glimpse-based approaches (\glimpse{}) on the foreground (see \cref{fig:output_examples_main}). 
However, if we consider the ability to segment the background (mIoU in~\cref{tab:results_clevr}), their performance advantage disappears, with SPAIR* performing the best.  
We attribute this to the tendency of pixel-space models to assign parts of the background to nearby objects. 
In glimpse-based methods, however, the formation of glimpses forces the objects to be spatially compact, which offers an advantage when separating the objects from the background. 

\paragraph{Textured Scenes} 
When training on \CLEVRTEX{}, all models struggle. 
The foreground segmentation performance reduces, indicating that models fail to assign whole objects to a single component, likely due to the tendency to overfit consistent color regions. 
The overall segmentation performance is worse as well. 
MSE is much higher than on \CLEVR, with models producing blurry or flat reconstructions, failing to capture much of the rich variation in the input data. 
SPAIR*, which showed the best overall performance on \CLEVR{}, fails to recognize any objects and instead simply predicts the background. 
We conjecture that SPAIR's autoregressive handling of objects paired with the use of spatial transformers might make the learning signal too noisy. 

Sprite-based models (\sprite{}) also perform worse, as the greater variation in appearances is not sufficiently captured by their limited dictionary. 
While the dictionary size can be increased, 
the lack of an internal compression mechanism to represent varied appearances will always be a limiting factor in natural world settings.
Interestingly, when unable to capture individual objects, MN learns to tile the image with possible color blobs, representing low-frequency information in the image instead. 
In our tests, similar tiling behavior tends to occur also in glimpse-based models whenever they cannot learn to reconstruct the foreground (see the Appendix,~\cref{fig:tiling_examples}, for examples in other models).
Since DTI includes a set of internal transformations, it performs comparatively better on \CLEVRTEX{}.

GNM, a generative glimpse-based approach, has overall the best performance on \CLEVRTEX{}, which we attribute to spatial-locality constraints imposed through the glimpse-based formulation and limited background reconstruction ability due to a simpler background model; \ie comparing to other methods less capacity is spent on the background.
Interestingly, GNM shows one of the largest reconstruction errors, despite being the best at scene segmentation, suggesting that ignoring confounding aspects of the scene rather than representing them might aid in the overall task.

Out of the our benchmarked pixel-space methods (\pixels{}), IODINE performs the best in terms of the overall segmentation performance. 
Our qualitative investigation shows that pixel-space methods that can segment textured scenes largely capture consistent color regions, which occasionally align with objects on scenes with simpler materials. 
Large patterns in the background or changes in object appearance, often due to lighting result in oversegmentation.

\paragraph{Stability}
Due to inherent stochasticity in initialization and optimization, one can expect a degree of variation between different model training runs. 
Many benchmarked models in this study also rely on internal randomness, primarily due to the sampling procedures involved. 
This influences the learning signal and the configuration the models can learn. 
Pixel-based approaches and SPACE (which has pixel-space model for background) show higher variance in the performance metrics. 
Similar to \citep{locatello2020object, emami2021efficient, lin2020space}, we observe that these methods occasionally fail to use separate components, which causes high fluctuation between different seeds. 
Glimpse-based methods are more stable with respect to seeds but tend to exhibit higher sensitivity to hyperparameter settings.

\paragraph{Generalisation}
In addition to benchmarking existing approaches in their ability to learn and handle textured scenes, we are also interested in the degree to which different approaches might rely on specific factors of \CLEVRTEX{}. 
To this end, we evaluate the models trained on the \CLEVRTEX{} on two additional test sets: \CAMO{} to see whether models rely on the difference of object appearances present in a scene, and \TEST{} to see whether a degree of memorization (\eg of shapes and materials) plays a role in recognition and whether the methods could generalize to unseen patterns.

Interestingly, some of the better performing approaches on \CLEVRTEX{} maintain much of their segmentation ability on out-of-distribution (\TEST{}) data. GNM, for example, attempts to reconstruct the input using memorized training data materials and shapes, which leads to reduced but still comparable object segmentation.
Other sprite- (\sprite{}) and glimpse-based (\glimpse{}) methods either do not perform well or show similar reliance on the appearances from the training distribution. 
Pixel-space models (\pixels{}) show a better ability to reconstruct the input but also tend to reconstruct based on consistent color regions rather than objects, a tendency only exacerbated by the out-of-distribution setting.

When considering the challenging \CAMO{} setting, none of the approaches perform satisfactory segmentation. 
Methods that somewhat work on \CLEVRTEX{} tend to use different components to represent lighter and darker parts of the scene, highlighting the tendency of all current models to overfit the scene appearance. 

\subsection{Variants}

\begin{figure}
    \includegraphics[width=.97\textwidth, trim={0 5 26 0}]{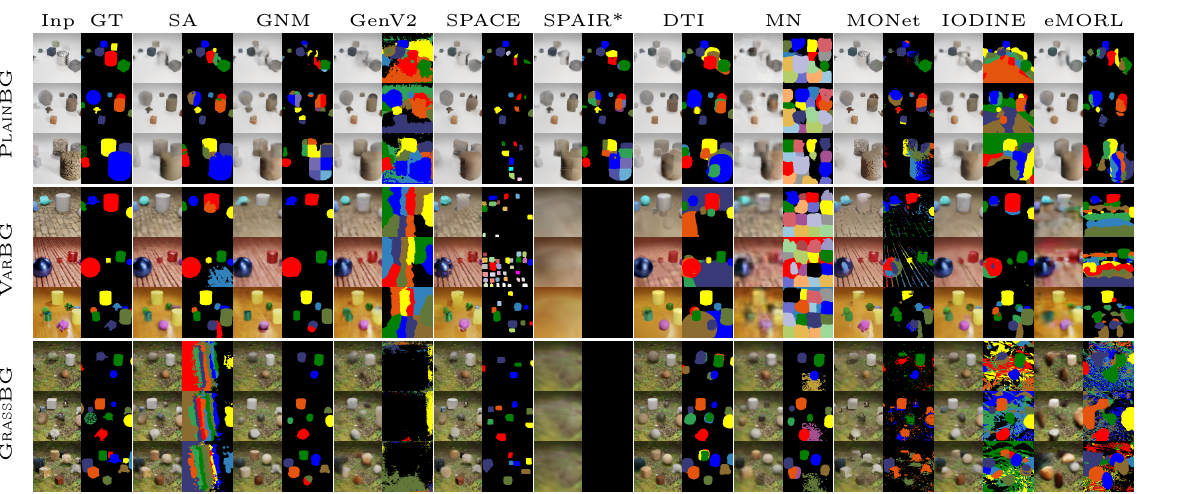}
	\caption{Comparison of various models' reconstruction and segmentation outputs on \PBG{}, \VBG{} and \GRASSBG{} variants. Best viewed digitally.}
	\label{fig:output_examples_variants}
\end{figure}

\begin{table}
\caption{Model results on \PBG{},\VBG, and \GRASSBG{} variants.}
\label{tab:results_variants}
\centering
\scalebox{.90}{
\begin{tabular}{lrrlrrlrr}
\toprule
\multirow{2}{*}{Model} & \multicolumn{2}{c}{\PBG{}} 
& 
      & \multicolumn{2}{c}{\VBG{}} 
&   
      & \multicolumn{2}{c}{\GRASSBG{}{}}
\\
& \(\uparrow\)mIoU (\%) & \(\downarrow\)MSE 
& 
& \(\uparrow\)mIoU (\%) & \(\downarrow\)MSE 
& 
& \(\uparrow\)mIoU (\%) & \(\downarrow\)MSE 
\\
  
\cmidrule[\lightrulewidth]{1-3}
\cmidrule[\lightrulewidth]{5-6}
\cmidrule[\lightrulewidth]{8-9}

\glimpse{} SPAIR* \citep{crawford2019spatially}
& 39.32 & 134
&
& 0.00 & 1246
&
& 0.00 & 728
\\

\glimpse{} SPACE \citep{lin2020space}
& 31.96 & 120
&
& 16.10 & 311
&
& 33.85 & 196
\\

\glimpse{} GNM \citep{jiang2020generative}
& 26.49 & 96
&
& 49.78 & 438
&
& 53.15 & 254
\\

\sprite{} MN \citep{smirnov2021marionette} 
& 10.16 & 167
&
& 11.51 & 441
&
& 34.80 & 266
\\

\sprite{} DTI \citep{monnier2021dtisprites} 
& 36.03 & 210
&
& 38.82 & 498
&
& 37.65 & 215
\\

\pixels{} GenV2 \citep{engelcke2021genesis} 
& 24.39 & 98
&
& 14.40 & 298
&
& 2.88 & 306
\\

\pixels{} eMORL \citep{emami2021efficient} 
& 29.39 & 96
&
& 22.92 & 385
&
& 19.38 & 199
\\

\pixels{} MONet \citep{burgess2019monet} 
& 38.72 & 83
&
& 23.73 & 212
&
& 21.29 & 165
\\

\pixels{} SA \citep{locatello2020object} 
& 39.32 & 134
&
& 62.57 & 257
&
& 12.88 & 116
\\

\pixels{} IODINE \citep{greff2019iodine} 
& 23.83 & 128
&
& 39.86 & 364
&
& 25.76 & 225
\\

\bottomrule
\end{tabular}
}
\end{table}

As discussed above, many of the models that perform well on \CLEVR{}, either do not work on \CLEVRTEX{} or lose much of their performance. 
To further probe which aspects of the scene composition are challenging, we use the variants of \CLEVRTEX{}.

\paragraph{Textured Objects}
When applied to \PBG{}, where materials are only seen on objects, and the background is gray, all of the methods still perform worse than on \CLEVR{}, with a significant drop in segmentation performance, especially prevalent in pixel-space approaches (\pixels{}).
Since all methods have been designed with simpler datasets and uniformly colored objects, the more realistic nature of the materials in \CLEVRTEX{} poses a difficult challenge. 
Glimpse-based models (\glimpse{}) also show reduced segmentation quality over \CLEVR{}. MN (sprite-based) struggles as the increased diversity in foreground objects overwhelms the spite dictionary.
Finally, the models' inability to capture the fine-grained details of the more complex object appearance causes the increase in reconstruction error.

\paragraph{Textured Background}
\VBG{} contains simple mono-colored objects arranged on top of a diverse set of textured backgrounds. Certain models, like SPAIR*, SPACE, and GenV2, struggle to handle diverse backgrounds. Other methods, however, seem to benefit from simpler objects, showing improvements in segmentation performance over both \PBG{} and \CLEVRTEX{} scenarios, indicating that these models rely on simpler, more consistent objects.

\paragraph{Consistent Background}
\GRASSBG{} has the same complex forest grass background in all scenes. 
The background is richer and more complex than in \PBG{}. 
As glimpse-space methods (\glimpse{}) tend to model the background explicitly, we observe that contrasting consistent background aids these models greatly. 
Pixel-space methods (\pixels{}) also perform slightly better in this setting than on \CLEVRTEX{} where the background varies. 
However, the effect is not as pronounced as for glimpse-based (\glimpse{}) approaches, with the overall performance roughly matching what was observed on \CLEVRTEX{}.

\section{Conclusions}

Unsupervised object learning and scene segmentation is a challenging task. 
Interestingly, given the existing metrics and commonly used datasets (\eg, \CLEVR{}), current approaches show impressive performance, yet we have shown that they are easily challenged when visual complexity increases. 
To this end, we present \CLEVRTEX{}, a new benchmark that aims to increase visual scene complexity, which contains richer textures, materials, and shapes, to encourage progress towards methods applicable to real images in the wild. 

In our experiments, GNM \cite{jiang2020generative} and IODINE \cite{greff2019iodine} perform the best out of glimpse-based and pixel-space models, respectively, with GNM showing the best segmentation performance overall. 
However, almost all methods struggle to handle multiple textured scenes, resulting in a significant performance gap with respect to the closest current benchmark, \CLEVR{}.
Our findings suggest that pixel-space methods tend to be more prone to overfitting consistent color regions and smooth gradients. 
On the other hand, sprite- and glimpse-based approaches tend to memorize small repeated patterns, which offers an advantage on \CLEVRTEX{}. Further testing, however, shows that these models reconstruct smooth backgrounds and recognize sharp changes as objects. 
As such, even the approaches that show some ability to handle textured environments focus largely on scene appearance, failing to learn and exploit global context clues that might align with semantic objects. 

We believe that textures pose a challenge to current pixel-space and glimpse-based methods as they are built to exploit simple visual elements and uniform appearance that is present in previous datasets, partly due to the reconstruction objectives. We find evidence for this in our experiments with the dataset variants: consistency within \textit{objects}, as seen in our \VBG{} variant, and consistency in \textit{backgrounds} (\PBG{} and \GRASSBG{}) helps to learn better models than the full \CLEVRTEX{} where there is no simple intra- and inter-appearance consistency.
Only on simpler scenes (\cref{fig:output_examples_main}) the best performing methods succeed at segmenting some objects.

Thus, \CLEVRTEX{} offers new challenges for unsupervised multi-object segmentation, especially for evaluating generalization. 
Furthermore, the three variants and two additional test sets can serve as a diagnostic tool for developing new methods, and the extensive evaluation acts as a standardized benchmark for current and future methods.

\paragraph{Limitations}
\label{sec:limitations}

The proposed dataset contains a limited number of primitive shapes and a catalog of 60 materials. 
Although future models might exploit the non-exhaustive nature of object appearance, \eg, memorizing object reconstructions than learning generalizable scene decompositions, we have shown that current methods are, in fact, faced with a significant challenge, even at a slight increase of data complexity (\eg, on \PBG{}).
To further address this limitation, we have created the \TEST{} dataset, which should serve as an additional test for the generalization ability of models outside the training distribution.  
Overall, \CLEVRTEX{} is still a synthetic dataset and does not fully close the gap to real-world data. However, until methods can solve \CLEVRTEX{}, generalization to real images is likely out of reach.

\paragraph{Broader Impact}
\label{sec:impact}
The work presented here critically evaluates current approaches for unsupervised multi-object segmentation. 
The introduced datasets are fully simulated renderings of 3D primitives and do not contain any people or personal information. 
Our benchmark aims to establish and standardize evaluation practices, provide new challenges for current algorithms, and help future research compare with prior work.
While \CLEVRTEX{} is highly important for current research, its impact outside of the research community is low as current methods can not yet properly deal with real images.

\begin{ack}
L.~K. is funded by EPSRC Centre for Doctoral Training in Autonomous Intelligent Machines and Systems EP/S024050/1. 
I.~L. is supported by the European Research Council (ERC) grant IDIU-638009 and EPSRC VisualAI EP/T028572/1.
C.~R. is supported by Innovate UK (project 71653) on behalf of UK Research and Innovation (UKRI) and by the ERC IDIU-638009. 
We thank \citet{johnson2017clevr} for their open-source implementation of \CLEVR{}. 
We would also like to thank Martin Engelcke for helpful suggestions on applying Genesis-V2 to \CLEVRTEX{}, Patrick Emami for assistance adapting eMORL to \CLEVRTEX{} and Dmitriy Smirnov for sharing their implementation of MarioNette.
\end{ack}

\section*{References}
\bibliographystyle{plainnat}
{
\small
\renewcommand{\bibsection}{}
\bibliography{references}
}
\newpage

\appendix
\section{Dataset Documentation: Datasheets for Datasets}\label{sec:datasheet}
Here we answer the questions outlined in the datasheets for datasets paper by \citet{gebru2018datasheets}.
\subsection{Motivation}
\paragraph{For what purpose was the dataset created?}
\CLEVRTEX{} was created to serve as the next challenging benchmark for unsupervised multi-object segmentation methods. It trades simpler visuals for confounding aspects such as texture, irregular shapes, and a variety of materials.

\paragraph{Who created the dataset (e.g., which team, research group) and on behalf of which entity (e.g., company, institution, organisation)?}
The dataset has been constructed by the research group ``Visual Geometry Group'' at the Engineering Science Department, University of Oxford.

\paragraph{Who funded the creation of the dataset?}
The dataset is created for research purposes at VGG. L. K. is funded by EPSRC Centre for Doctoral Training in Autonomous Intelligent Machines and Systems EP/S024050/1. I. L. is supported by the EPSRC programme grant Seebibyte EP/M013774/1 and ERC starting grant IDIU-638009. C. R. is supported by Innovate UK (project 71653) on behalf of UK Research and Innovation (UKRI) and by the European Research Council (ERC) IDIU-638009.

\subsection{Composition}
\paragraph{What do the instances that comprise the dataset represent (e.g., documents, photos, people, countries)?}
The dataset consists of images featuring simulated scenes and segmentation, depth, normal, albedo, and shadow masks available, and metadata detailing scene composition.

\paragraph{How many instances are there in total (of each type, if appropriate)?}
There are $50\,000$ instances in the main \CLEVRTEX{} dataset. $20\,000$ in each variant, \PBG{}, \VBG{}, \GRASSBG{} and \CAMO{}. There is also a further $10\,000$ instances in the testing-only variant \TEST{}.

\paragraph{Does the dataset contain all possible instances or is it a sample (not necessarily random) of instances from a larger set?}
The dataset is a sample of the near-infinite set of possible arrangements under our sampling distribution. Please see \cref{sec:creation} for a description of the process to sample the scene. 

\paragraph{What data does each instance consist of?}
Each instance consists of the RGB scene image, depth, normal, albedo, and shadow masks (all PNG), and further metadata (JSON) detailing object positions, shapes, scales, and materials used. We use only the RBG image for training during the benchmarking process and segmentation masks and metadata to evaluate.

\paragraph{Is there a label or target associated with each instance?}
For the task explored in this paper, unsupervised multi-object segmentation, the target labels are the segmentation masks, which are not used during training.

\paragraph{Is any information missing from individual instances?}
No.

\paragraph{Are relationships between individual instances made explicit (e.g., users’ movie ratings, social network links)?}
No, there are no relationships between different instances. 

\paragraph{Are there recommended data splits (e.g., training, development/validation, testing)?}
Yes, we adopt 10\%/10\%/80\% test/val/train splits for the datasets by instance index, with the exception of \TEST{} variant, which is used for evaluation only. The rationale behind splits is that the data comes from the same generation process for each variant and can already be considering randomized. Simply using an image index to separate the splits makes both data-loading easy and removes the need to distribute canonical split indexes.

\paragraph{Are there any errors, sources of noise, or redundancies in the dataset?} %
No. 

\paragraph{Is the dataset self-contained, or does it link to or otherwise rely on external resources (e.g., websites, tweets, other datasets)?}
The dataset is self-contained.

\paragraph{Does the dataset contain data that might be considered confidential (e.g., data that is protected by legal privilege or by doctor-patient confidentiality, data that includes the content of individuals’ non-public communications)?}
No.

\paragraph{Does the dataset contain data that, if viewed directly, might be offensive, insulting, threatening, or might otherwise cause anxiety?}
No.

\paragraph{Does the dataset relate to people? If not, you may skip the remaining questions in this section.}
No.

\paragraph{Does the dataset identify any subpopulations (e.g., by age, gender)?} NA

\paragraph{Is it possible to identify individuals (i.e., one or more natural persons), either directly or indirectly (i.e., in combination with other data) from the dataset?} NA

\paragraph{Does the dataset contain data that might be considered sensitive in any way (e.g., data that reveals racial or ethnic origins, sexual orientations, religious beliefs, political opinions or union memberships, or locations; financial or health data; biometric or genetic data; forms of government identification, such as social security numbers; criminal history)?} NA

\subsection{Collection process}
\paragraph{How was the data associated with each instance acquired?}
The data was generated.

\paragraph{What mechanisms or procedures were used to collect the data (e.g., hardware apparatus or sensor, manual human curation, software program, software API)?}
The images were rendered using Blender 2.9.3 software on generic systems.

\paragraph{If the dataset is a sample from a larger set, what was the sampling strategy (e.g., deterministic, probabilistic with specific sampling probabilities)?}
See the similar question in the Composition section.

\paragraph{Who was involved in the data collection process (e.g., students, crowdworkers, contractors) and how were they compensated (e.g., how much were crowdworkers paid)?}
The authors were involved in the process of generating this dataset.

\paragraph{Over what timeframe was the data collected?} 
The datasets were rendered over a period of several weeks.

\paragraph{Were any ethical review processes conducted (e.g., by an institutional review board)?}
No.

\paragraph{Does the dataset relate to people? If not, you may skip the remainder of the questions in this section.}
No.

\subsection{Preprocessing/cleaning/labeling}
\paragraph{Was any preprocessing/cleaning/labeling of the data done (e.g., discretization or bucketing, tokenization, part-of-speech tagging, SIFT feature extraction, removal of instances, processing of missing values)?}
No, the dataset was generated together with labels.

\paragraph{Was the “raw” data saved in addition to the preprocessed/cleaned/labeled data (e.g., to support unanticipated future uses)?} %
NA

\paragraph{Is the software used to preprocess/clean/label the instances available?}
NA

\subsection{Uses}
\paragraph{Has the dataset been used for any tasks already?}
In the paper we show and benchmark the intended use of this dataset for unsupervised multi-object segmentation setting.

\paragraph{Is there a repository that links to any or all papers or systems that use the dataset?}
We will be listing these on the website.

\paragraph{What (other) tasks could the dataset be used for?}
We include additional information maps when generating this dataset, which could be used for exploring value of using extra modalities for supervision or as targets. As mentioned before, we also generated necessary metadata for CLEVR-like QA task.

\paragraph{Is there anything about the composition of the dataset or the way it was collected and preprocessed/cleaned/labeled that might impact future uses?}
No.

\paragraph{Are there tasks for which the dataset should not be used?}
This dataset is meant for research purposes only.

\subsection{Distribution}
\paragraph{Will the dataset be distributed to third parties outside of the entity (e.g., company, institution, organization) on behalf of which the dataset was created?}
No. 

\paragraph{How will the dataset will be distributed (e.g., tarball on website, API, GitHub)?}
The dataset and related evaluation code is available on the website \url{https://www.robots.ox.ac.uk/~vgg/research/clevrtex/} allowing users to download and read-in the data. 

\paragraph{When will the dataset be distributed?}
The dataset is available now.

\paragraph{Will the dataset be distributed under a copyright or other intellectual property (IP) license, and/or under applicable terms of use (ToU)?}
CC-BY.

\paragraph{Have any third parties imposed IP-based or other restrictions on the data associated with the instances?}
The original textures used in rendering objects are copyrighted by Poliigon Pty Ltd and cannot be redistributed to a third party. This only applies to texture images used in creating this dataset. The materials used for main dataset are freely available under non-commercial license and we include instructions to retrieve them alongside the generation code. Textures used in evaluation-only \TEST{} variant are not available free of charge (we obtained them under a commercial license), but their catalogue is similarly included with the code. The dataset instances themselves do not have IP-based restrictions.

\paragraph{Do any export controls or other regulatory restrictions apply to the dataset or to individual instances?}
Not that we are are of. Regular UK laws apply. 

\subsection{Maintenance}
\paragraph{Who is supporting/hosting/maintaining the dataset?}
The dataset is supported by the authors and by the VGG research group. 
The main contact person is Laurynas Karazija.

\paragraph{How can the owner/curator/manager of the dataset be contacted (e.g., email address)?}
The authors of this dataset can be reached at their e-mail addresses: \textit{\{laurynas,chrisr,iro\}@robots.ox.ac.uk}.

\paragraph{Is there an erratum?}
If errors are found and erratum will be added to the website.

\paragraph{Will the dataset be updated (e.g., to correct labeling errors, add new instances, delete instances)?}
Any potential future updates or extension will be communicated via the website. The dataset will be versioned.

\paragraph{If the dataset relates to people, are there applicable limits on the retention of the data associated with the instances (e.g., were individuals in question told that their data would be retained for a fixed period of time and then deleted)?}
NA
\paragraph{Will older versions of the dataset continue to be supported/hosted/maintained?}
We plant to continue hosting older versions of the dataset.

\paragraph{If others want to extend/augment/build on/contribute to the dataset, is there a mechanism for them to do so?}
Yes, we make the dataset generation code available.

\subsection{Other questions}

\paragraph{Is your dataset free of biases?}
Yes.

\paragraph{Can you guarantee compliance to GDPR?}
No, we are unable to comment on legal issues.

\subsection{Author statement of responsibility} 
The authors confirm all responsibility in case of violation of rights and confirm the licence associated with the dataset and its images.  

\section{Dataset}
The dataset can be accessed at \url{https://www.robots.ox.ac.uk/~vgg/research/clevrtex}. In \CLEVRTEX{} and its variants, each instance contains:
\begin{enumerate}
\item RBG scene image
\item semantic mask image
\item depth mask image
\item shadow mask image
\item albedo mask image
\item normal mask image
\item Metadata JSON, which further details:
    \begin{enumerate}
    \item number of objects
    \item background material
    \item shape of each object
    \item size of each object
    \item rotation of each object
    \item scene (3D) coordinates of each object
    \item image (2D) coordinates of each object
    \item material of each object
    \item color (only relevant on \VBG{}) of each object
    \item scene directions (\CLEVR{} metadata)
    \item object relationships (\CLEVR{} metadata)
    \end{enumerate}
\end{enumerate}
All images are provided as PNG. We also provide code for reading in the dataset and evaluation utilities for general performance metrics and per-shape/material/size breakdown. The dataset is provided under the CC-BY license.

\section{Supplementary Material}
\subsection{Data}
\label{sec:sup_data}
All images are center-cropped to a $192 \times 192$ patch and further downsampled to $128 \times 128$ pixels as a pre-processing step before being fed to the models.
This introduces partially visible objects in the datasets, removes uninteresting empty edges of the scenes, and lowers the computational load. 
Many of the benchmarked models were developed to work with such resolution.
We include helper code to load our datasets for convenience.
For \CLEVR{} we are using a version that includes segmentation masks for evaluation\footnote{Available at \texttt{\url{https://github.com/deepmind/multi_object_datasets}}.}, for which we adopt the standard 70k/15k/15k train/validation/test splits. 

\subsection{Metrics}
\label{sec:sup_metrics}
As previously mentioned, prior work \citep{greff2019iodine, engelcke2019genesis, locatello2020object} evaluated using the adjusted Rand index (ARI) metric calculated only on pixels that correspond to the foreground objects, filtered using ground-truth data. We share the concern of some authors \citep{engelcke2019genesis, monnier2021dtisprites} that such evaluation protocol does not account for whether objects are considered a part of the background and how well models segment object boundaries. 
Instead, we opt for the mIoU metric, familiar from the supervised segmentation setting. The predicted objects are matched with ground truth segments using the Hungarian matching algorithm, which assigns only a single predicted component to each true mask, maximizing overall overlap. A mean is taken over all objects, including the background. We provide side-by-side comparison of these metrics on all benchmarked models in \cref{tab:results_cle_clt_aris,tab:results_var_aris}. We chose mIoU in favor of ARI metric, as it weights all objects equally irrespective of their size. ARI is based on counting pairs, thus it gives larger regions such backgrounds more weight. 

\newcommand{\arif}[2]{
\hspace*{-.5em}\tablenum[table-format = 2.2]{#1}\small\color{DarkSlateBlue}$\pm$\tablenum[table-format = 2.2]{#2}
}
\newcommand{\arifp}[1]{
\hspace*{-.5em}\tablenum[table-format = 2.2]{#1}\hspace*{3.1em}
}

\newcommand{\mious}[2]{
\hspace*{-1.2em}\tablenum[table-format = 2.2]{#1}\small\color{DarkSlateBlue}$\pm$\tablenum[table-format = 2.2]{#2}\hspace*{-1.3em}
}
\newcommand{\mioups}[1]{
\hspace*{-1.2em}\tablenum[table-format = 2.2]{#1}\hspace*{2.4em}
}
\begin{table}
\caption{Benchmark results on \CLEVR{}, \CLEVRTEX{}, \CAMO{}, and \TEST{} comparing ARI-FG and mIoU metrics. Results are shown $(\pm \sigma)$ calculated over 3 runs.}
\label{tab:results_cle_clt_aris}
\scalebox{.64}{
\begin{tabular}{lrrlrrlrrlrr}
\toprule
\multirow{2}{*}{Model} & \multicolumn{2}{c}{\CLEVR{}} 
& 
      & \multicolumn{2}{c}{\CLEVRTEX{}} 
&   
      & \multicolumn{2}{c}{\TEST{}}
&   
      & \multicolumn{2}{c}{\CAMO{}}
\\
& \(\uparrow\)ARI-FG (\%) 
& \(\uparrow\)mIoU (\%) 
& 
& \(\uparrow\)ARI-FG (\%) 
& \(\uparrow\)mIoU (\%) 
& 
& \(\uparrow\)ARI-FG (\%) 
& \(\uparrow\)mIoU (\%) 
&
& \(\uparrow\)ARI-FG (\%) 
& \(\uparrow\)mIoU (\%) 
\\
\cmidrule[\lightrulewidth]{1-3}
\cmidrule[\lightrulewidth]{5-6}
\cmidrule[\lightrulewidth]{8-9}
\cmidrule[\lightrulewidth]{11-12}

\glimpse{} SPAIR* \citep{crawford2019spatially}
& \arif{77.13}{1.92} & \mious{65.95}{4.02} 
&
& \arif{0.00}{0.00} & \mious{0.00}{0.00} 
&
& \arif{0.00}{0.00} & \mious{0.00}{0.00} 
&
& \arif{0.00}{0.00} & \mious{0.00}{0.00} 
\\

\glimpse{} SPACE \citep{lin2020space} 
& \arif{22.75}{14.04} & \mious{26.31}{12.93} 
&
& \arif{17.53}{4.13} & \mious{9.14}{3.46} 
&
& \arif{12.71}{3.44} & \mious{6.87}{3.32} 
&
& \arif{10.55}{2.09} & \mious{8.67}{3.50} 
\\

\glimpse{} GNM \citep{jiang2020generative} 
& \arif{65.05}{4.19} & \mious{59.92}{3.72}
&
& \arif{53.37}{0.67} & \mious{42.25}{0.18} 
&
& \arif{48.43}{0.86} & \mious{40.84}{0.30} 
&
& \arif{15.73}{0.89} & \mious{17.56}{0.74} 
\\

\sprite{} MN \citep{smirnov2021marionette} 
& \arif{72.12}{0.64} & \mious{56.81}{0.40} 
&
& \arif{38.31}{0.70} & \mious{10.46}{0.10} 
&
& \arif{37.29}{1.04} & \mious{12.13}{0.19} 
&
& \arif{31.52}{0.87} & \mious{8.79}{0.15} 
\\

\sprite{} DTI \citep{monnier2021dtisprites} 
& \arif{89.54}{1.44} & \mious{48.74}{2.17}
&
& \arif{79.90}{1.37} & \mious{33.79}{1.30} 
&
& \arif{73.67}{0.98} & \mious{32.55}{1.08} 
&
& \arif{72.90}{1.89} & \mious{27.54}{1.55} 
\\

\pixels{} GenV2 \citep{engelcke2021genesis} 
& \arif{57.90}{20.38} & \mious{9.48}{0.55} 
&
& \arif{31.19}{12.41} & \mious{7.93}{1.53} 
&
& \arif{29.04}{11.23} & \mious{8.74}{1.64}
&
& \arif{29.60}{12.84} & \mious{7.49}{1.67}
\\

\pixels{} eMORL \citep{emami2021efficient} 
& \arif{93.25}{3.24} & \mious{50.19}{22.56}
&
& \arif{45.00}{7.77} & \mious{12.58}{2.39}
&
& \arif{43.13}{9.28} & \mious{13.17}{2.58}
&
& \arif{42.34}{7.19} & \mious{11.56}{2.09}
\\

\pixels{} MONet \citep{burgess2019monet} 
& \arif{54.47}{11.41} & \mious{30.66}{14.87}
&
& \arif{36.66}{0.87} & \mious{19.78}{1.02}
&
& \arif{32.97}{1.00} & \mious{19.30}{0.37}
&
& \arif{12.44}{0.73} & \mious{10.52}{0.38}
\\

\pixels{} SA \citep{locatello2020object} 
& \arif{95.89}{2.37} & \mious{36.61}{24.83} 
&
& \arif{62.40}{2.23} & \mious{22.58}{2.07} 
&
& \arif{58.45}{1.87} & \mious{20.98}{1.59} 
&
& \arif{57.54}{1.01} & \mious{19.83}{1.41} 
\\

\pixels{} IODINE \citep{greff2019iodine} 
& \arif{93.81}{0.76} & \mious{45.14}{17.85}
&
& \arif{59.52}{2.20} & \mious{29.17}{0.75}
&
& \arif{53.20}{2.55} & \mious{26.28}{0.85}
&
& \arif{36.31}{2.57} & \mious{17.52}{0.75}
\\

\bottomrule
\end{tabular}
}
\end{table}
\begin{table}
\caption{Results on \PBG{},\VBG, and \GRASSBG{} variants, comparing ARI-FG and mIoU metrics.}
\label{tab:results_var_aris}
\centering
\scalebox{0.65}{
\begin{tabular}{lrrlrrlrr}
\toprule
\multirow{2}{*}{Model} & \multicolumn{2}{c}{\PBG{}} 
& 
      & \multicolumn{2}{c}{\VBG{}} 
&   
      & \multicolumn{2}{c}{\GRASSBG{}{}}
\\
& \(\uparrow\)ARI-FG (\%) & \(\uparrow\)mIoU (\%) 
& 
& \(\uparrow\)ARI-FG (\%) & \(\uparrow\)mIoU (\%) 
& 
& \(\uparrow\)ARI-FG (\%) & \(\uparrow\)mIoU (\%) 
\\
  
\cmidrule[\lightrulewidth]{1-3}
\cmidrule[\lightrulewidth]{5-6}
\cmidrule[\lightrulewidth]{8-9}

\glimpse{} SPAIR* \citep{crawford2019spatially}
& 51.75 & 39.32 
&
& 0.05 & 0.00 
&
& 0.00 & 0.00 
\\

\glimpse{} SPACE \citep{lin2020space} 
& 34.25 & 31.96 
&
& 29.36 & 16.10 
&
& 32.52 & 33.85 
\\

\glimpse{} GNM \citep{jiang2020generative} 
& 40.73 & 26.49 
&
& 66.79 & 49.78 
&
& 67.31 & 53.15 
\\

\sprite{} MN \citep{smirnov2021marionette} 
& 38.34 & 10.16 
&
& 43.64 & 11.51 
&
& 59.79 & 34.80 
\\

\sprite{} DTI \citep{monnier2021dtisprites} 
& 77.74 & 36.03 
&
& 81.56 & 38.82 
&
& 82.37 & 37.65 
\\

\pixels{} GenV2 \citep{engelcke2021genesis} 
& 85.33 & 24.39 
&
& 66.04 & 14.40 
&
& 21.12 & 2.88 
\\

\pixels{} eMORL \citep{emami2021efficient} 
& 52.00 & 29.39
&
& 50.18 & 22.92 
&
& 69.64 & 19.38
\\

\pixels{} MONet \citep{burgess2019monet} 
& 57.10 & 38.72
&
& 51.87 & 23.73 
&
& 37.97 & 21.29 
\\

\pixels{} SA \citep{locatello2020object} 
& 51.75 & 39.32 
&
& 89.78 & 62.57 
&
& 43.55 & 12.88 
\\

\pixels{} IODINE \citep{greff2019iodine} 
& 54.32 & 23.83 
&
& 75.33 & 39.86 
&
& 66.91 & 25.76 
\\

\bottomrule
\end{tabular}
}
\end{table}

\subsection{Hyper-parameters}
\label{sec:hyperparameters}
Where available in PyTorch, we use the official implementation for the benchmarked methods. Otherwise, we use a re-implementation, checked against the original method, and further verify that it produces similar results to those reported in the corresponding papers. 
Where the original methods have been applied to \CLEVR{} (or its variant), we employ the same hyper-parameter configuration for \CLEVR{}. 
For other datasets or methods that have not been trained on \CLEVR{}, we follow a best-effort approach to tuning hyper-parameters. 

For MONet~\cite{burgess2019monet}, we reduced the batch size from 64 to 63 (\(3 \times 21\)). IODINE~ \cite{greff2019iodine} and MONet were trained for 300k iterations instead of 1M as we noticed that no changes to learned configurations, running loss, or performance improvements were taking place after 250k iterations. For MONet, IODINE we found the original configuration worked well enough. 
For SPACE~\cite{lin2020space}, we concentrated on finding a suitable setting for output standard deviation for foreground and background networks. Despite higher values being crucial for both Genesis and GNM models, we could not identify a configuration that produced better results than the original 0.15 in our exploration. The following describes any adjustments made to the original configurations for other models.

\paragraph{Slot Attention \cite{locatello2020object}} We use 11 slots on all tests. We varied the number of attention iterations. We have found the model to perform the best when trained using 3. We maintained the original learning rate, batch size, and optimizer settings and trained for the suggested 500k iterations.

\paragraph{Efficient MORL \citep{emami2021efficient}} We increase the number of components to 11 and change the input resolution to \(128 \times 128\) to be inline with other methods studied. GECO reconstruction target is further adjusted to account for change in resolution. We use the value of $-108\,000$ for \CLEVR{} and \PBG{}. We use higher values of $-61\,000$ for \VBG{} and \GRASSBG{} and $-73\,000$ for \CLEVRTEX{}, \TEST{}, and \CAMO{}, due to more complex backgrounds. We considered a set of $\{-8\,000, -48\,000, -61\,000, -69\,000, -73\,000, -108\,000, -112\,000\}$, selecting the best performing ones. eMORL$^\dagger$: following the release of \CLEVRTEX{}, the codebase of eMORL has been updated including configuration settings for \CLEVRTEX{}. The authors provided us with trained models that show better performance (\cref{tab:results_clevr}) in our evaluation.

\paragraph{GNM \cite{jiang2020generative}} We use a \(4 \times 4\) slot grid with total of 16 slots and a latent dimension of 64 for objects and 10 for background. We found the model extremely sensitive to the output standard deviation. We found values 0.2 on \CLEVR{} and 0.5 on \CLEVRTEX{} worked well. It is worth noting that in our testing, with values of 0.4 and 0.6, GNM could not learn to segment the scene. We trained for 300k iteration.

\paragraph{GenesisV2 \cite{engelcke2021genesis}} We focused our hyper-parameter selection on the output standard deviation and GECO \citep{rezende2018taming} objective. On \CLEVR{} we used GECO goal of 0.5655 and output standard deviation of 0.7, which was crucial for model to learn as lower values did not produce good segmentations. On \CLEVRTEX{} we lowered the GECO goal to 0.5, which outperformed \CLEVR{} setting. 

\begin{wraptable}{r}{.45\textwidth}
\vspace{-1.2em}
\caption{Architecture of component networks changed in SPAIR*.}
\center
\scalebox{.75}{
\begin{tabular}{lccc}
\multicolumn{4}{l}{\textbf{Conv Encoder}} \\
\toprule
Layer & Size/Ch. & Act. & Comment \\
\midrule
Conv \(3 \times 3\) & 32 & ReLU & stride 2 \\
Conv \(3 \times 3\) & 32 & ReLU & stride 2 \\
Conv \(3 \times 3\) & 64 & ReLU & stride 2 \\
Conv \(3 \times 3\) & 64 & ReLU & stride 2 \\
Avg P \(1 \times1\) & & & \\
MLP & 128 & ReLU & \\
MLP & \(||\mu|| + ||\sigma||\) & \multicolumn{2}{l}{Softplus for \(\sigma\) only} \\
\bottomrule
\\
\multicolumn{4}{l}{\textbf{Broadcast Decoder}} \\
\toprule
Layer & Size/Ch. & Act. & Comment \\
\midrule
Broadcast & & & add coord. \\
Conv \(3 \times 3\) & 32 & ReLU & no pad \\
Conv \(3 \times 3\) & 32 & ReLU & no pad \\
Conv \(3 \times 3\) & 32 & ReLU & no pad \\
Conv \(3 \times 3\) & 32 & ReLU & no pad \\
Conv \(1 \times 1\) & 4 & \multicolumn{2}{l}{Sigmoid for masks only}  \\
\bottomrule
\end{tabular}
}
\vspace{-1em}
\end{wraptable}
\paragraph{SPAIR* \cite{crawford2019spatially}} As mentioned before, we incorporated a background VAE network into SPAIR by using a convolutional encoder and a spatial broadcast decoder \cite{watters2019spatial}. We also replaced MLP-based glimpse decoder with a similar spatial broadcast decoder. Additionally, we added an extra convolution in the backbone network to handle inputs of \(128 \times 128\) size. In this configuration, SPAIR had 16 slots. We set the latent dimension of objects to 64, and background to 1 on \CLEVR{} and 4 on \CLEVRTEX{}. We trained for 250k iterations using a batch size of 128, Adam optimizer, learning rate of 1e-4, with gradient clipping when norm exceeded 1.0. We used \(\beta\) value of 2.7. On \CLEVR{} we used the output standard deviation of 0.15. On \CLEVRTEX{}, we annealed the value from 0.5 to 0.15 over 50k iterations. On \CLEVR{}, the object presence prior hyper-parameter \(s\) was annealed from 0.0001 to 0.99 over 10k, on \CLEVRTEX{}, over 50k iterations.

\paragraph{DTISprites \cite{monnier2021dtisprites}} On \CLEVR{}, we used the setting used for CLEVR6 in the original work except for increasing the possible number of objects. We found that using ten slots leads to better segmentation results than setting to 11 as with other models (one more than the max number of objects). On \CLEVRTEX{}, we used color and protective transforms for both sprites and backgrounds.

\paragraph{MarioNette \cite{smirnov2021marionette}} We adjusted the model to learn to select and use from a dictionary of backgrounds, same as sprites. Additionally, we lowered the layer size to 4, using two layers, which gives 32 possible slots of size \(64 \times 64\). On \CLEVRTEX{}, we increased the sizes of both background and sprite dictionaries to as large as would fit in GPU memory.  We trained with 60 sprites and single background on \CLEVR{}, \PBG{}, and \GRASSBG{} increasing the number of backgrounds to 60 on \VBG{} and \CLEVRTEX{}.

\subsection{Extra Figures}

\begin{figure}[!h]
    \includegraphics[width=.99\textwidth, trim={0 5 26 0}]{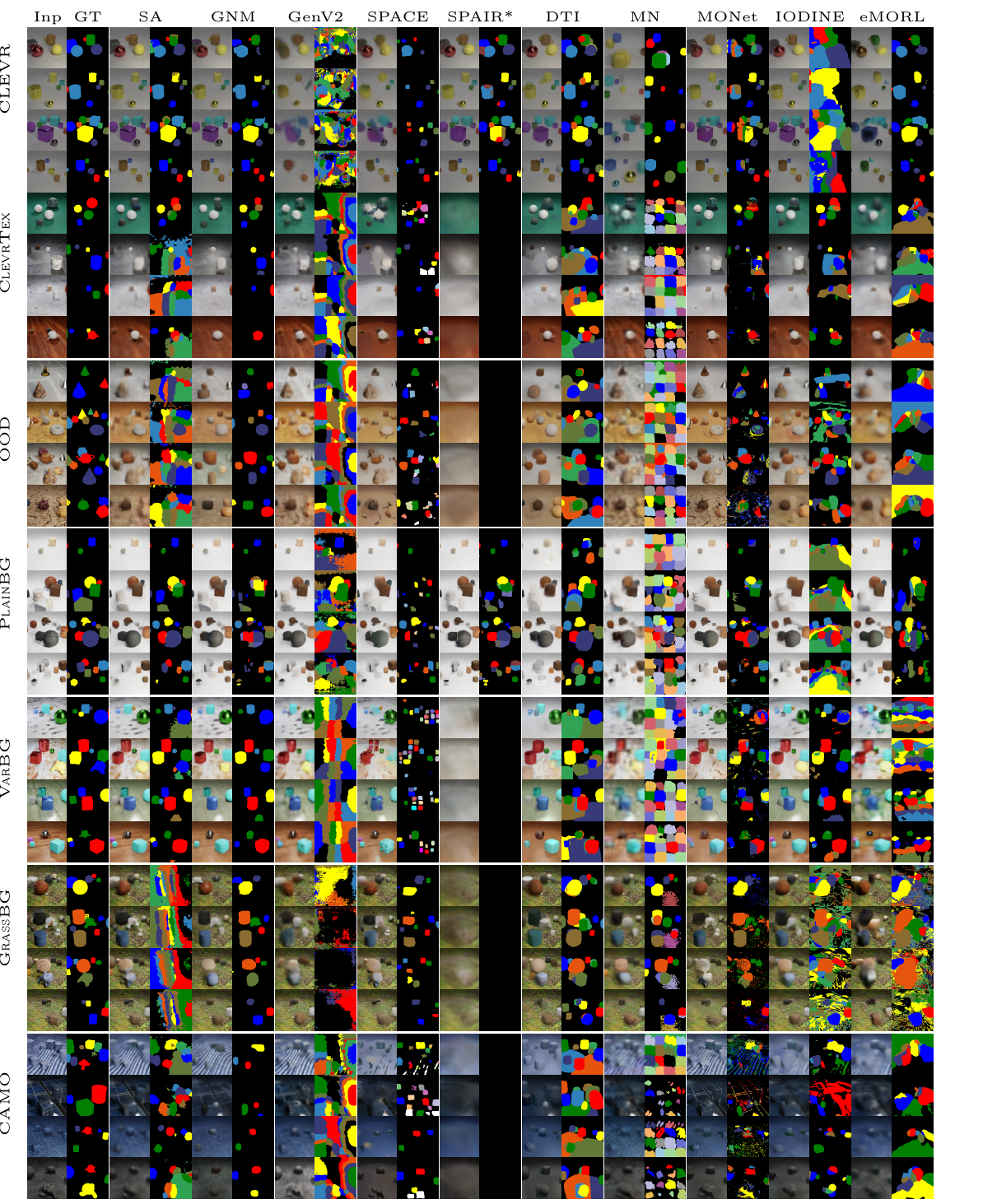}
	\caption{Comparison of various model reconstruction and segmentation outputs on \CLEVR{}, \CLEVRTEX{} and variants. Best viewed digitally.}
	\label{fig:output_examples}
\end{figure}

Here, we include extra figures listing additional output for all benchmarked models on \CLEVR{}, \CLEVRTEX{}, test sets and variants (\cref{fig:output_examples}). \cref{fig:tiling_examples} contains example output of sprite- and glimpse-based models when they fail to learn correct foreground and background elements and learn to tile the image instead.

\newcommand{\tilingoutputaddimage}[1]{
    \includegraphics[width=\textwidth, trim={4 4 4 4}]{#1}
}
\begin{figure}[t!]
    \centering
    \vspace*{-0.3em}
    \begin{subfigure}[tl]{0.245\textwidth}
		\begin{subfigure}[l]{0.49\textwidth}
		    \caption*{\\Input}
			\tilingoutputaddimage{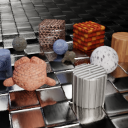}
			\tilingoutputaddimage{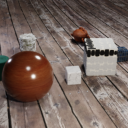}
			\tilingoutputaddimage{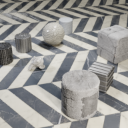}
			\tilingoutputaddimage{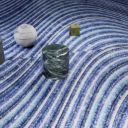}
		\end{subfigure}
		\begin{subfigure}[l]{0.49\textwidth}
		    \caption*{Ground\\Truth}
			\tilingoutputaddimage{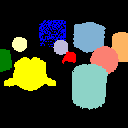}
			\tilingoutputaddimage{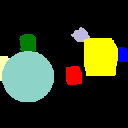}
			\tilingoutputaddimage{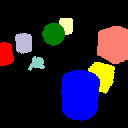}
			\tilingoutputaddimage{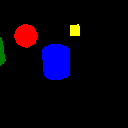}
		\end{subfigure}
	\end{subfigure}
	\begin{subfigure}[tl]{0.245\textwidth}
		\caption*{\\MN~\cite{smirnov2021marionette}}
		\begin{subfigure}[l]{0.49\textwidth}
			\tilingoutputaddimage{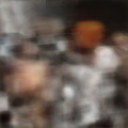}
			\tilingoutputaddimage{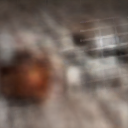}
			\tilingoutputaddimage{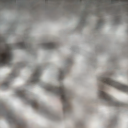}
			\tilingoutputaddimage{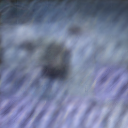}
		\end{subfigure}
		\begin{subfigure}[l]{0.49\textwidth}
			\tilingoutputaddimage{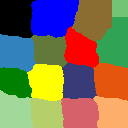}
			\tilingoutputaddimage{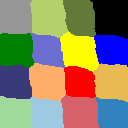}
			\tilingoutputaddimage{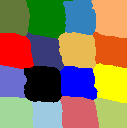}
			\tilingoutputaddimage{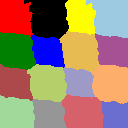}
		\end{subfigure}
	\end{subfigure}
    \begin{subfigure}[tl]{0.245\textwidth}
		\caption*{\\GNM~\cite{jiang2020generative}}
		\begin{subfigure}[l]{0.49\textwidth}
			\tilingoutputaddimage{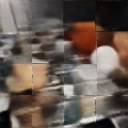}
			\tilingoutputaddimage{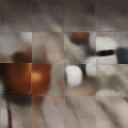}
			\tilingoutputaddimage{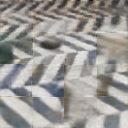}
			\tilingoutputaddimage{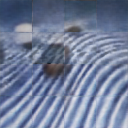}
		\end{subfigure}
		\begin{subfigure}[l]{0.49\textwidth}
			\tilingoutputaddimage{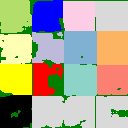}
			\tilingoutputaddimage{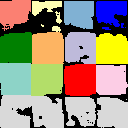}
			\tilingoutputaddimage{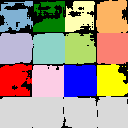}
			\tilingoutputaddimage{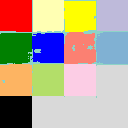}
		\end{subfigure}
	\end{subfigure}
	\begin{subfigure}[tl]{0.245\textwidth}
		\caption*{\\SPACE~\cite{lin2020space}}
		\begin{subfigure}[l]{0.49\textwidth}
			\tilingoutputaddimage{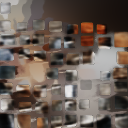}
			\tilingoutputaddimage{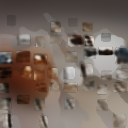}
			\tilingoutputaddimage{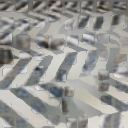}
			\tilingoutputaddimage{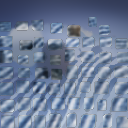}
		\end{subfigure}
		\begin{subfigure}[l]{0.49\textwidth}
			\tilingoutputaddimage{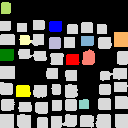}
			\tilingoutputaddimage{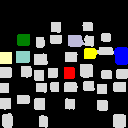}
			\tilingoutputaddimage{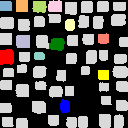}
			\tilingoutputaddimage{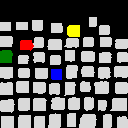}
		\end{subfigure}
	\end{subfigure}
	\vspace*{-0.8em}

    \caption{Tiling behaviour common to glimpse- (\glimpse{}) and sprite-based (\sprite{}) models. Such tiling occurred whenever the model could not reproduce the foreground and background elements with respective component networks to sufficient accuracy. The models are trained on \CLEVRTEX{}. GNM is shown here trained with output \(\sigma = 0.3\).}
    \label{fig:tiling_examples}
\end{figure}

\subsection{Dataset Construction}
\label{sec:dataset_construction}

\begin{figure}[t]
    \begin{subfigure}[l]{0.33\textwidth}
        \includegraphics[width=\textwidth]{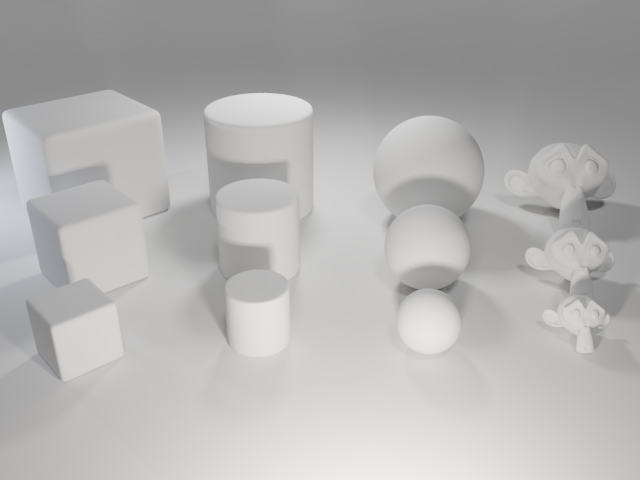}
    \end{subfigure}
    \begin{subfigure}[l]{0.33\textwidth}
        \includegraphics[width=\textwidth]{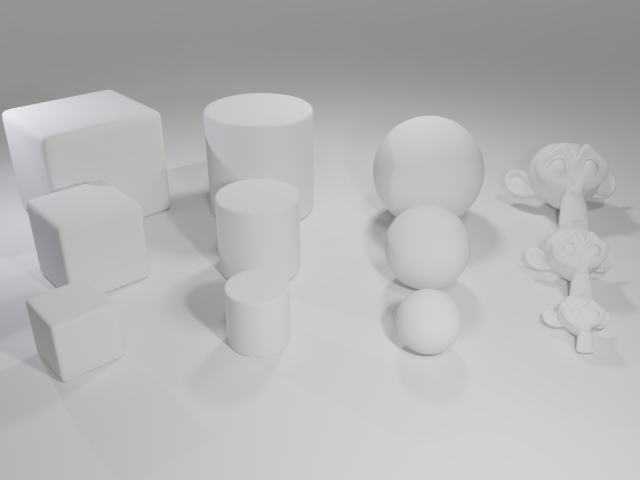}
    \end{subfigure}
    \begin{subfigure}[l]{0.33\textwidth}
        \includegraphics[width=\textwidth]{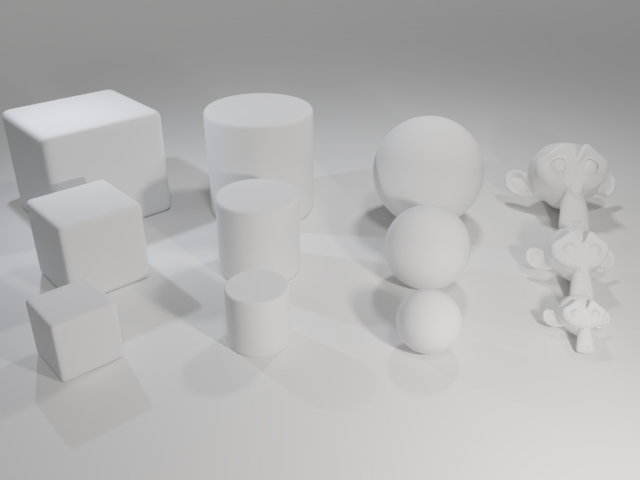}
    \end{subfigure}
	\caption{Effects of jittering light positions in the scenes. The images show two extremes with the mean position in the middle. The images also contain a showcase of 4 shapes present in the main \CLEVRTEX{} dataset at 3 possible scales. The scenes are rendered without any materials.}
	\label{fig:light_jitter}
\end{figure}

The main method of how the dataset is constructed is described in \cref{sec:creation}. Here, we include additional figures to showcase some steps in the dataset creation and provide catalog of materials used.
\paragraph{Lighting} \cref{fig:light_jitter} shows the possible range of randomizing light positions in the scene, from warm closeup light positions with lots of shadows falling onto other objects to distant lights casting small soft shadows onto background even in crowded scenes. \cref{fig:light_jitter} also shows 4 possible shapes at 3 possible scales used in the \CLEVRTEX{}.

\begin{figure}
    \centering
    \includegraphics[width=.7\textwidth]{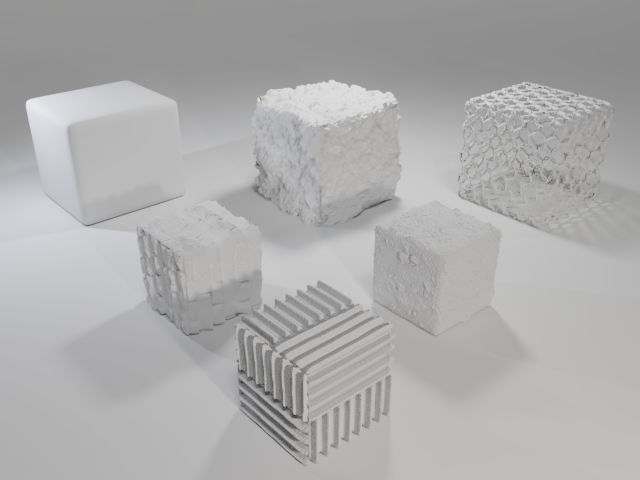}
	\caption{Showcase of a diverse set of shape perturbations applied the basic cube (top left) through a combination of displacement mapping, bumping and transparency mapping. Other material properties are not applied to the objects to show only the displacement details.}
	\label{fig:displacement}
\end{figure}

\paragraph{Shape Adjustments} \CLEVRTEX{} features only 4 simple objects. This is mitigated by a range of material-specific geometry adjustments, bumping and transparency mapping applied to the seed shapes. \cref{fig:displacement} shows the effect of the shape perturbations in a scene where no other material properties have been applied to the objects.

\paragraph{Camera} The camera position is jitterred along with lights. We use a perspective camera with a focal length of 0.035m and 0 shift.

\begin{figure}[t]
    \begin{subfigure}[l]{0.49\textwidth}
        \includegraphics[width=\textwidth]{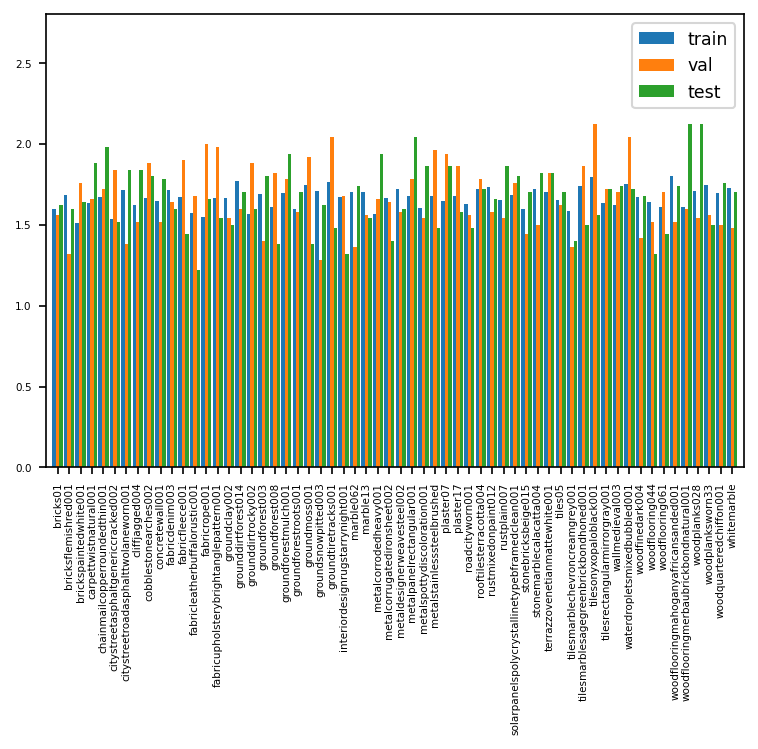}
        \caption{Background materials}
    \end{subfigure}
    \begin{subfigure}[l]{0.49\textwidth}
        \includegraphics[width=\textwidth]{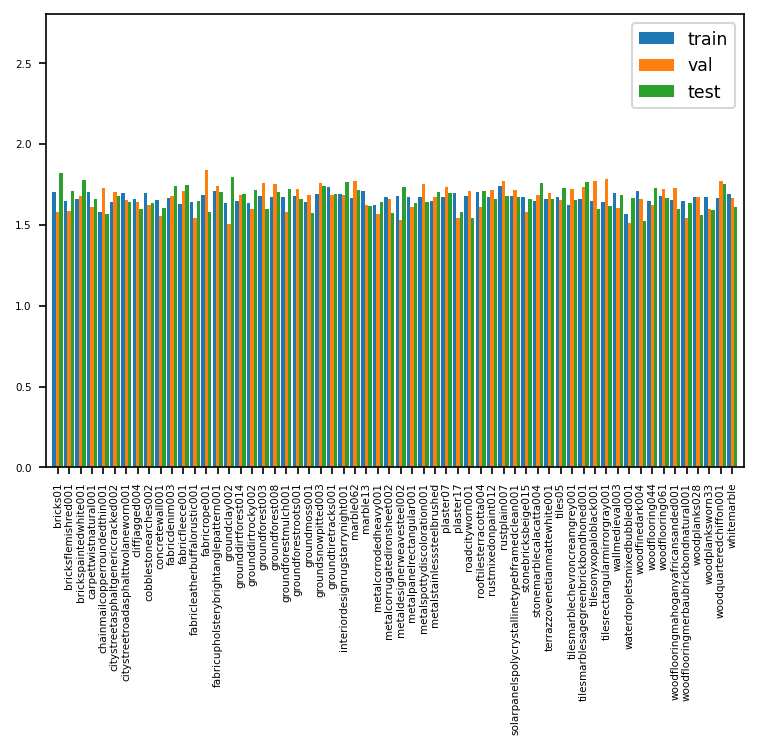}
        \caption{Object materials}
    \end{subfigure}
	\caption{Distribution of 60 materials in \CLEVRTEX{} dataset between train/val/test splits, shown as a percentage. (a) shows distribution for the background. (b) shows distribution for objects.}
	\label{fig:mats_barchart}
\end{figure}
\paragraph{Dataset Splits} \CLEVRTEX{} and variants are split into test/val/train datasets using 10\%/10\%/80\% proportions after generation. The splits are made based on the index of the example, that is first 10\% form test split. This simple scheme is motivated by the uniform sampling of the scene composition. \cref{fig:mats_barchart} shows that this results in roughly proportional distribution of materials for both backgrounds and objects across dataset splits. \TEST{} variant is test-only.

\begin{figure}
    \centering
    \includegraphics[width=.9\textwidth]{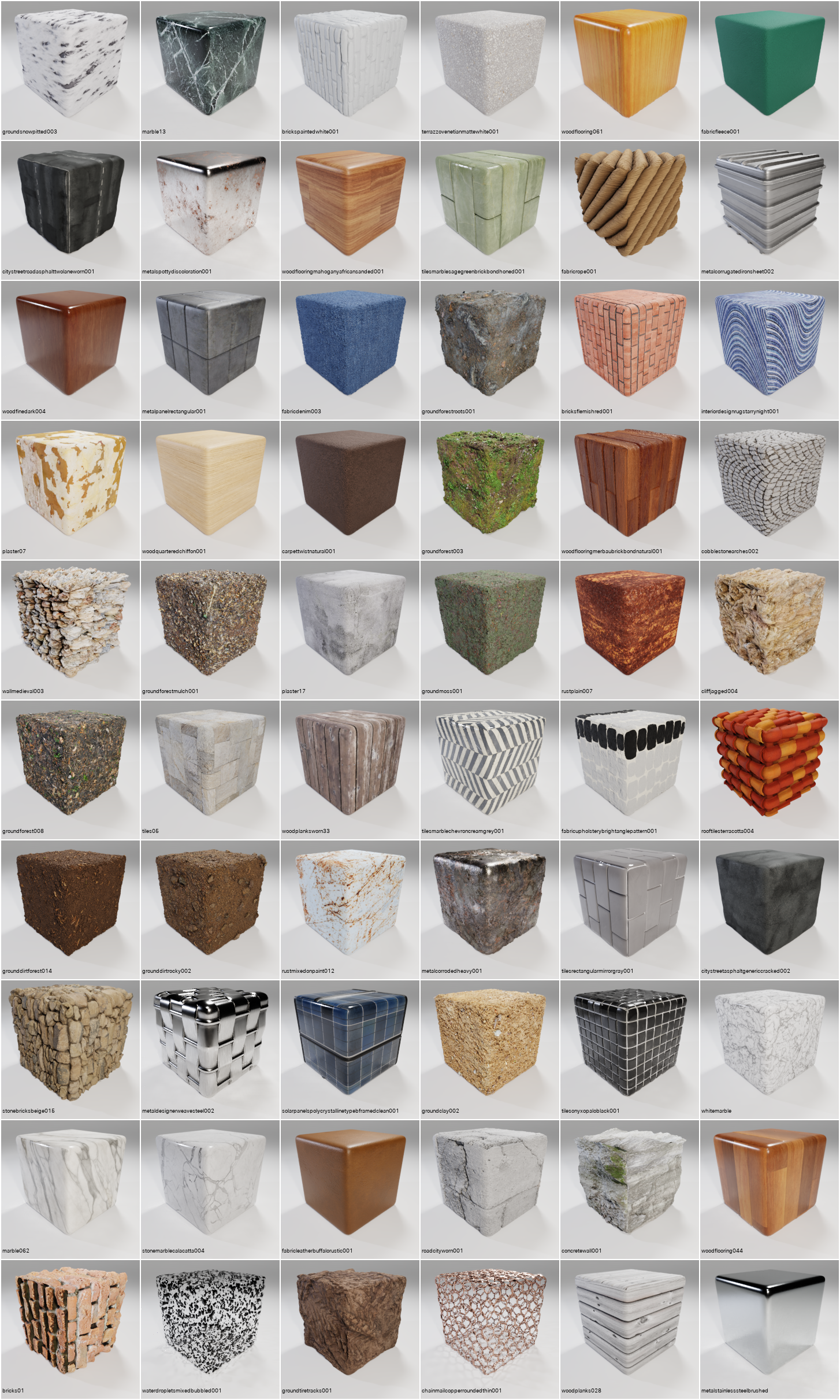}
	\caption{Materials used in \CLEVRTEX{} dataset.}
	\label{fig:mats_clevrtex}
\end{figure}

\begin{figure}
    \centering
    \includegraphics[width=.9\textwidth]{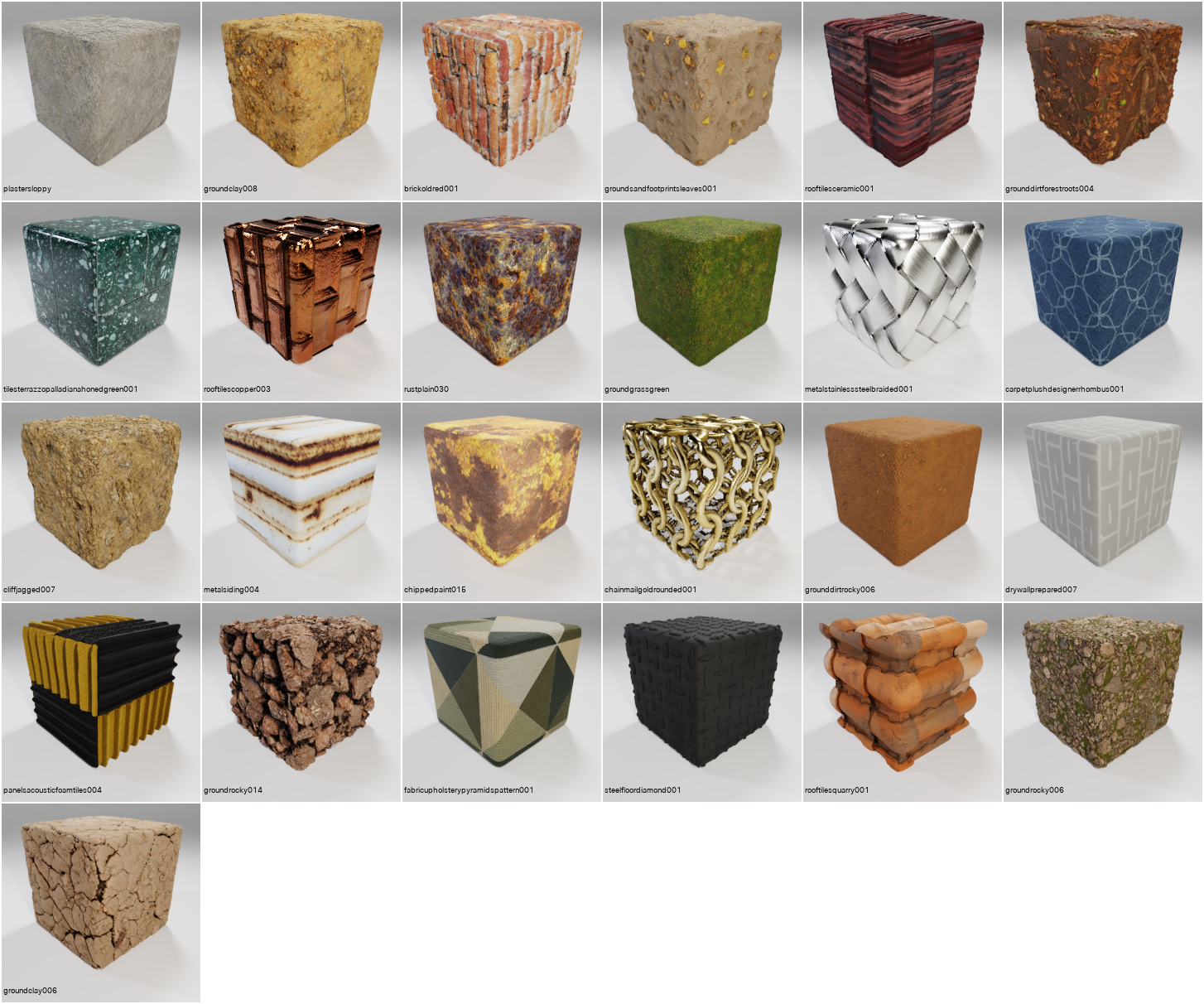}
	\caption{Materials used in \TEST{} dataset variant.}
	\label{fig:mats_ood}
\end{figure}

\paragraph{Materials} \cref{fig:mats_clevrtex} contains the list of 60 materials used in generating \CLEVRTEX{} and its \PBG{}, \VBG{}, \GRASSBG{}, and \CAMO{} variants. Please see our generation code for further information. \cref{fig:mats_ood} contains 25 materials used in \TEST{} variant.

\end{document}